%% file: main.tex
\documentclass{article}

\usepackage{microtype}
\usepackage{graphicx}
\usepackage{subfigure}
\usepackage{booktabs} % for professional tables

\usepackage{hyperref}

\usepackage[accepted]{icml2025}

\usepackage{amsmath}
\usepackage{amssymb}
\usepackage{mathtools}
\usepackage{amsthm}
\usepackage{hyperref}
\usepackage{url}
\usepackage{amsmath}
\usepackage{amssymb}
\usepackage{graphicx}
\usepackage{multicol}
\usepackage{xspace}
\usepackage{color}
\usepackage{caption}
\usepackage{comment}
\usepackage{tcolorbox}
\usepackage{listings}
\usepackage{alltt}
\usepackage{multirow}
\usepackage{float}

\usepackage[capitalize,noabbrev]{cleveref}

\usepackage{subcaption}

\theoremstyle{plain}

\theoremstyle{definition}

\theoremstyle{remark}

\usepackage[textsize=tiny]{todonotes}
\usepackage{titlesec}
\titlespacing{\section}{0pt}{10pt}{5pt}
\titlespacing{\subsection}{0pt}{8pt}{4pt}
\titlespacing{\subsubsection}{0pt}{6pt}{3pt}

\icmltitlerunning{LAST SToP For Modeling Asynchronous Time Series}

\begin{document}

\twocolumn[
\icmltitle{LAST SToP For Modeling Asynchronous Time Series}



\icmlsetsymbol{equal}{*}

\begin{icmlauthorlist}
\icmlauthor{Shubham Gupta}{sch}
\icmlauthor{Thibaut Durand}{comp}
\icmlauthor{Graham Taylor}{yyy}
\icmlauthor{Lilian W. Bia\l okozowicz}{comp}
\end{icmlauthorlist}

\icmlaffiliation{yyy}{Vector Institute}
\icmlaffiliation{comp}{RBC Borealis}
\icmlaffiliation{sch}{Université Laval}

\icmlcorrespondingauthor{Shubham Guptss}{shgup1@ulaval.ca}
\icmlcorrespondingauthor{Thibaut Durand}{thibaut.durand@borealisai.com}
\icmlcorrespondingauthor{Graham Taylor}{gwtaylor@uoguelph.ca}
\icmlcorrespondingauthor{Lilian W. Bia\l okozowicz}{lilian.wong@borealisai.com}

\icmlkeywords{Machine Learning, ICML}

\vskip 0.3in
]



\printAffiliationsAndNotice{}  % leave blank if no need to mention equal contribution

\begin{abstract}
We present a novel prompt design for Large Language Models (LLMs) tailored to \textbf{Asynchronous Time Series}. Unlike regular time series, which assume values at evenly spaced time points, asynchronous time series consist of timestamped events occurring at irregular intervals, each described in natural language. Our approach effectively utilizes the rich natural language of event descriptions, allowing LLMs to benefit from their broad world knowledge for reasoning across different domains and tasks. This allows us to extend the scope of asynchronous time series analysis beyond forecasting to include tasks like anomaly detection and data imputation.

We further introduce \textbf{Stochastic Soft Prompting}, a novel prompt-tuning mechanism that significantly improves model performance, outperforming existing fine-tuning methods such as QLoRA. Through extensive experiments on real-world datasets, we demonstrate that our approach achieves state-of-the-art performance across different tasks and datasets.
\end{abstract}

\input{section/intro}

\input{section/sota}

\input{section/method}

\input{section/exp}
\input{section/conclusion}

\bibliography{section/refs}
\bibliographystyle{icml2025}

%%%%%%%%%%%%%%%%%%%%%%%%%%%%%%%%%%%%%%%%%%%%%%%%%%%%%%%%%%%%%%%%%%%%%%%%%%%%%%%
%%%%%%%%%%%%%%%%%%%%%%%%%%%%%%%%%%%%%%%%%%%%%%%%%%%%%%%%%%%%%%%%%%%%%%%%%%%%%%%
% APPENDIX
%%%%%%%%%%%%%%%%%%%%%%%%%%%%%%%%%%%%%%%%%%%%%%%%%%%%%%%%%%%%%%%%%%%%%%%%%%%%%%%
%%%%%%%%%%%%%%%%%%%%%%%%%%%%%%%%%%%%%%%%%%%%%%%%%%%%%%%%%%%%%%%%%%%%%%%%%%%%%%%
\newpage
\appendix
\onecolumn
\input{section/appendix}

\end{document}

%% file: section/intro.tex
\section{Introduction}

An asynchronous time series (also named \textit{temporal event sequence} or \textit{continuous-time event sequence}) is a temporally ordered set of events that describe the progression of actions or occurrences. Asynchronous time series are ubiquitous in daily life, such as healthcare \citep{Lorch2018Stochastic, Rizoiu2018SirHawkes}, finance \citep{Bacry2015Hawkes, Jin2020Visual}, e-commerce \citep{Hernandez2017Analysis}, and social media \citep{Zhang2022Counterfactual, Kong2023Interval}. In each of those domains, predicting the next events plays a crucial role. %Unlike time series which carry regular time stamps, asynchronous time series data is a sequence of events that do not necessarily follow any time pattern and modeling them has presented new challenges.%  

Unlike regular time series, which consist of values at evenly spaced time intervals (like weather measurements), asynchronous time series consist of multiple types of discrete events occurring sporadically over time. For example, in the context of social media platforms like X (Twitter), user interactions (likes, comments, shares, and follows) happen sporadically and at irregular intervals~\citep{zhao2015seismic}.  Each such type of interaction with a user's profile represents an event type, and together with their timestamps, form an asynchronous time series~ \citep{Xue2023EasyTPP}. Modeling such asynchronous time series is challenging due to the irregular timing and the diversity of event types, which contrasts with the uniformity and regularity of traditional time series data~\citep{schirmer2022modeling,horn2020set,zhangirregular}.

% {Unlike regular time series, which consist of numerical values at evenly spaced time intervals, asynchronous time series involve events occurring at irregular intervals without any consistent time pattern. This irregularity, combined with the complexity of event descriptions, makes modeling asynchronous time series significantly more challenging than regular time series. \cite{schirmer2022modeling} \citep{horn2020set} \citep{zhangirregular} }

Traditionally, to model asynchronous time series, events are grouped into a fixed, small number of categorical types~\citep{Xue2023EasyTPP}. Separate stochastic processes—such as Poisson processes or Hawkes processes—are then modeled for each event type to predict which event will occur next and when~\citep{Mei2022Transformer,Hawkes1971Spectra}. However, this approach presents several significant drawbacks. \textit{Firstly}, it inherently limits research to datasets with a small number of event types because modeling each event type separately becomes increasingly computationally intensive as the number of event types grows \citep{Zuo2020Transformer}. \textit{Secondly}, events can vary widely and may not fit neatly into predefined categories. \textit{Thirdly}, this method leads to the loss of meaningful natural language descriptions associated with the events. \textit{Fourthly}, these methods treat each event type independently, ignoring any interactions between them --- for example, likes and shares of a tweet are not independent events. \textit{Lastly}, extending these methods to other tasks require significant theoretical development~ \citep{shchur2021detecting}. % Therefore, there is a need for modeling approaches that can handle a larger number of event types and leverage the detailed information available in natural language event descriptions. }

Deep learning models have significantly revolutionized techniques for time series modeling, and even more so with the introduction of transformers \citep{Vaswani2017Attention}. However, there are often limitations due to the scarcity of training data, overfitting in specific domains, and the highly specialized architectural designs. In response to those challenges, Large Language Models (LLMs) have emerged as a powerful and promising direction to model time series data. For example,  \citet{Gruver2023Large, Zhou2023One, Xue2023Promptcast, Jin2023TimeLLM} have illustrated how LLMs can be used as time series forecasters when the input time series is encoded as a string of numeric digits, by  casting the time series forecasting problem as a next-token prediction in text, hence unlocking the use of powerful pre-trained models. LLMs have also been explored in other domains like action forecasting from videos \citep{zhaoantgpt, wang2024lifelongmemory}. {However, these approaches focus on regular time series with evenly spaced numerical observations and cannot be directly applied to asynchronous time series due to their irregular intervals and diverse event types described in natural language. While LLMs have recently been explored for action recognition and action forecasting from videos \citep{zhaoantgpt,wang2024lifelongmemory}, applying LLMs to textual asynchronous time series over multiple tasks (like anomaly detection and imputation) remains largely unexplored. }

% Inspired by the successes of pre-trained language foundation models, we asked whether LLMs could be adapted to work on asynchronous time series data.

% \begin{figure}[ht]
% % \centering
% % \begin{subfigure}[b]{0.48\textwidth}
% %     \includegraphics[width=\textwidth]{figures/next_token_prediction}
% %     \caption{\textbf{Next token prediction.} The model is given a sequence of words/tokens with the goal of predicting the next word/token.}
% %     \label{fig:intro_next_token}
% % \end{subfigure}

\begin{figure}[ht]
\centering
\begin{minipage}[b]{\columnwidth}
\centering
    \includegraphics[width=\linewidth]{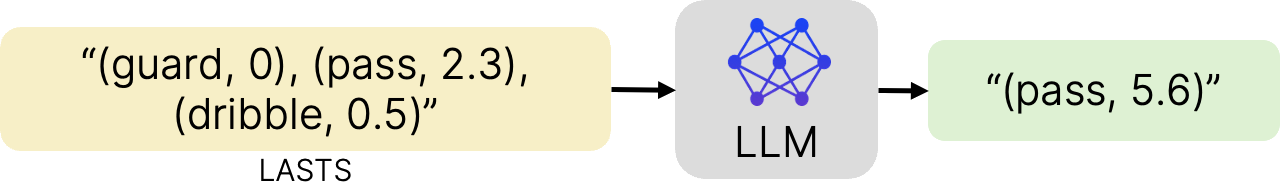}
    % \caption{\textbf{Forecasting} The model is given a sequence of events, encoded as text, with the goal of predicting the next event.}
    % \caption{\textbf{Forecasting}}
    % \textbf{(a) Forecasting}
    \label{fig:intro_forecasting}
\end{minipage} 
\begin{minipage}[b]{\columnwidth}
\centering
    \includegraphics[width=\linewidth]{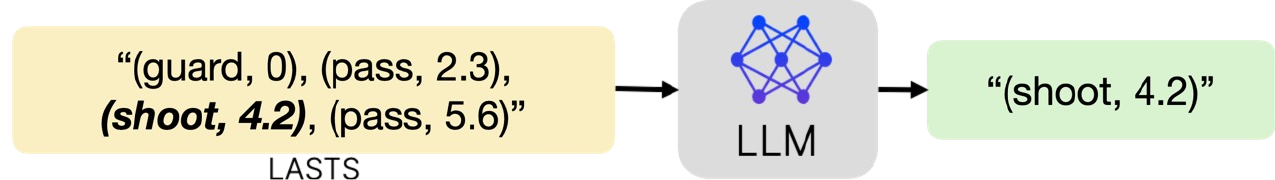}
    % \caption{\textbf{Anomaly detection} The model is given a sequence of events containing an incorrect event (bold) with the goal of finding the incorrect event.}
    % \textbf{(b) Anomaly detection}
    \label{fig:intro_anomaly_detection}
\end{minipage}
\begin{minipage}[b]{\columnwidth}
\centering
    \includegraphics[width=\linewidth]{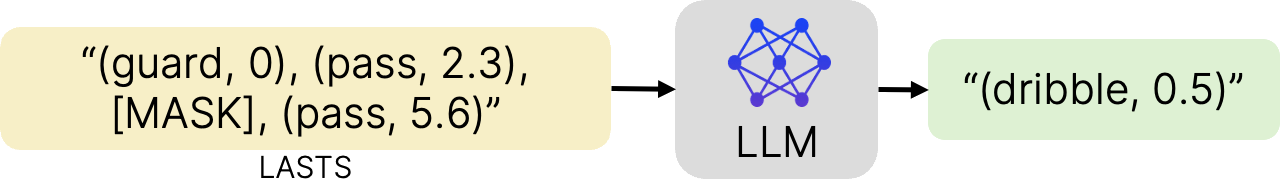}
    % \caption{\textbf{Imputation} The model is given a sequence of events containing a masked event, encoded as text, with the goal of predicting the masked event.}
    % \caption{\textbf{Imputation}}
    % \textbf{(c) Imputation}
    \label{fig:intro_imputation}
\end{minipage}
\caption{We show that our LASTS framework can solve the following tasks on asynchronous time series data: (a)\textbf{Forecasting:} \textit{(top)} The model is given a sequence of events, encoded as text, with the goal of predicting the next event. (b)\textbf{Anomaly detection:} \textit{(middle)} The model is given a sequence of events containing an incorrect event (bold) with the goal of finding the incorrect event. (c)\textbf{Imputation: }\textit{(bottom)} The model is given a sequence of events containing a masked event, encoded as text, with the goal of predicting the masked event.}

\label{fig:overall}
\end{figure}

%\caption{\textbf{LASTS}, the framework we propose for using LLMs to model asynchronous time series by encoding the sequence of events with natural language labels. LLMs have shown great performances to solve NLP tasks by predicting the next token given a sequence of tokens. We show that our model can solve tasks on asynchronous time series data such as  forecasting, anomaly detection and imputation. \lilian{I made the caption more concise.}}

% \caption{We propose a new method, named LASTS, for modeling asynchronous time series with LLMs, by encoding the sequence of events as text.
% (a) LLMs have shown great performances to solve NLP tasks by predicting the next token given a sequence of tokens. We show this property can be used to solve tasks on asynchronous time series data such as forecasting (b), anomaly detection (c) and imputation (d). 
% Each event is represented by the inter-arrival time of its occurrence and its event type. Unlike standard asynchronous time series models, our approach uses natural language event descriptions to extract richer semantic representations, resulting in greater accuracy.}

This paper presents \textbf{LASTS} (\textbf{L}anguage-modeled-\textbf{As}ynchronous \textbf{T}ime \textbf{S}eries), a novel prompting-based framework to adapt LLMs to asynchronous time series data while keeping the backbone model intact. To the best of our knowledge, this is the first work to explore the capabilities of LLMs to process textual asynchronous time series data and works on multiple tasks as shown in \autoref{fig:overall}. {Our framework overcomes the drawbacks presented by traditional approaches for modeling asynchronous time series --- it can handle datasets with numerous event types easily, it does not need to group events into predefined categorical bundles, it retains and utilizes the natural language descriptions of event types, and it is able to leverage the rich interactions between different event types.} Our contributions can be summarized as follows: 

\begin{itemize}
    \item \textbf{We introduce LASTS (Language-modeled Asynchronous Time Series)}, a novel framework that leverages Large Language Models (LLMs) to model asynchronous time series data, while effectively handling datasets with a large number of event types without the need for predefined categorical groupings. To the best of our knowledge, this is the first work to explore the capabilities of LLMs to process textual asynchronous time series data across multiple tasks such as forecasting, anomaly detection, and data imputation.

\item \textbf{We introduce Stochastic Soft Prompting (StoP)} which is an novel prompt-tuning mechanism that serves as a parameter-efficient method to adapt LLMs to asynchronous time series data. StoP learns soft prompts that significantly improve model performance and generalizability by randomly truncating the prompts during training to learn more diverse representations.

% \item \textbf{LASTS extends the scope of asynchronous time series analysis} beyond forecasting. Our framework allows the application of LLMs to tasks like anomaly detection and data imputation in asynchronous time series, which were previously unexplored in this domain, thus demonstrating the versatility of our approach.

\item \textbf{We conduct comprehensive evaluations} on real-world datasets across multiple tasks to demonstrate the effectiveness of our proposed method. Our approach achieves state-of-the-art performance, outperforming existing methods, and highlights the potential of LLM-based models to effectively process and analyze asynchronous time series data.
\end{itemize}

%% file: section/sota.tex
\section{Related Work}

\paragraph{Temporal Point Processes (TPPs).} 
TPPs \citep{Hawkes1971Spectra, daley2007introduction} have emerged as the standard method to model asynchronous time series data. 
% It is a mathematical framework for modeling sequences of events  and autoregressively generate events one after another. 
Over the last decade, a large number of neural temporal point processes have been proposed to capture complex dynamics of stochastic processes in time by using neural networks.
\citet{duTPP, MeiEisner} proposed to use models based on Recurrent Neural Networks (RNNs) to model the sequence of events. Then, more advanced models \citep{Mehrasa2019Variational, DiffusionTPP} were proposed to better model uncertainty when predicting the future. Recently, several neural TPP models incorporate Transformers in order to improve performance by using attention to better model long-term dependencies. These include the Self-attentive Hawkes process (SAHP) \citep{Zhang2020Self}, Transformer Hawkes process (THP) \citep{Zuo2020Transformer}, and Attentive Neural Hawkes Process (Att-NHP) \citep{Mei2022Transformer}.

\paragraph{Transformers for Time Series.} 
Transformers \citep{Vaswani2017Attention} have become popular to model regularly-sampled time series because of their ability to capture long-range dependencies and to extract semantic correlations among the elements of a long sequence. Informer \citep{Zhou2021Informer} introduced a novel self-attention architecture to reduce the quadratic complexity of the original self-attention. Autoformer \citep{Wu2021Autoformer} used a novel decomposition architecture with an auto-correlation mechanism to identify more reliable temporal patterns.  Crossformer \citep{Zhang2023Crossformer} proposed a novel architecture to model both  the cross-time and cross-dimension dependencies multivariate time series forecasting. PatchTST \citep{Yuqi2023PatchTST} tokenizes the time series in patches, and proposes a channel-independent patch time series Transformer to improve the long-term forecasting accuracy.

Due to space limitations, we only review some popular models and invite the reader to \citep{Wen2023Transformers, Zeng2023Transformers} for a more complete literature reviews of Transformer models for regularly-sampled time series.
Most of the time series Transformer models are designed for specific tasks, and cannot be easily extended to asynchronous time series data or other tasks like anomaly detection or imputation.

% Most of these Transformer models are designed for the forecasting task, and cannot be easily extended to other tasks like anomaly detection or imputation.
% Similarly, these models are designed for regularly-sampled time series and cannot be easily extended to asynchronous time series due to the different nature of the data.

\paragraph{Foundation Models (FMs) for Time Series.}
FMs \citep{Bommasani2021Opportunities} are a family of deep models that are pretrained on vast amounts of data, and have caused a paradigm shift due to their unprecedented capabilities for zero-shot and few-shot generalization.
FMs have revolutionized natural language processing \citep{Brown2020Language, BigScience2023Bloom, Wu2024Empirical, Dubey2024Llama3} and computer vision \citep{Radford2021Learning, Kirillov2023Segment}.
% Recently, several methods have been developed to build FMs for time series and can be organized in three main categories based on the pretraining data: text, time series, and image.
The availability of large-scale time series datasets has opened the door to pretrain a large model on time series data.
ForecastPFN \citep{Dooley2024ForecastPFN} proposed the first zero-shot forecasting method trained purely on synthetic data.
Lag-Llama \citep{Rasul2023LagLlama} introduced a univariate probabilistic forecasting model that was pretrained on a large corpus of diverse time series data.
TimeFM \citep{Das2024TimesFM} pretrained a decoder style attention model with input patching, using a large time series corpus comprising both real-world and synthetic
datasets. Chronos \citep{ansari2024chronos} introduced a framework for pretraining on tokenized time series data, achieving state-of-the-art zero-shot forecasting performance and simplifying forecasting workflows.
MOIRAI \citep{Woo2024Unified} is an enhanced Transformer architecture pretrained in the Large-scale Open Time Series Archive, that achieves competitive performance as a zero-shot forecaster.
% \subparagraph{Text-based FMs.} LLMs pretrained on large amounts of text data have emerged as a promising direction to model time series data.
% GPT4TS \citep{Zhou2023One}, LLM4TS \citep{Chang2024Llm4ts}, and TEMPO \citep{Cao2024Tempo} fine-tuned a pretrained GPT2 \citep{Radford2019Language} on some time series downstream tasks to capture intrinsic dynamic properties.
% TimeLLM \citep{Jin2023TimeLLM} proposed a reprogramming framework to repurpose LLMs for general time series forecasting with the backbone language models kept intact.
% PromptCast \citep{Xue2023Promptcast} introduced  a new prompt-based forecasting paradigm, where the numerical input and output are transformed into prompts and the forecasting task is framed in a sentence-to-sentence manner.
% LLMTime \citep{Gruver2023Large} showed that LLMs can zero-shot extrapolate time series if the numerical values of the time series are well represented.
\paragraph{{LLMs for Time Series}} LLMs pretrained on large amounts of text data have emerged as a promising direction to model time series data.
GPT4TS \citep{Zhou2023One}, LLM4TS \citep{Chang2024Llm4ts}, and TEMPO \citep{Cao2024Tempo} fine-tuned a pretrained GPT2 \citep{Radford2019Language} on some time series downstream tasks to capture intrinsic dynamic properties.
TimeLLM \citep{Jin2023TimeLLM} proposed a reprogramming framework to repurpose LLMs for general time series forecasting with the backbone language models kept intact.
PromptCast \citep{Xue2023Promptcast} introduced  a new prompt-based forecasting paradigm, where the numerical input and output are transformed into prompts and the forecasting task is framed in a sentence-to-sentence manner.
LLMTime \citep{Gruver2023Large} showed that LLMs can zero-shot extrapolate time series if the numerical values of the time series are well represented. LLM Processes \citep{requeima2024llm} explores various prompt configurations for using LLMs for time series forecasting condiitoned on a textual context. We refer the reader to \citep{zhang2024large} for a more detailed survey on the topic.

% \subparagraph{Time series-based FMs.} The availability of large-scale time series datasets has opened the door to pretrain a large model on time series data.
% ForecastPFN \citep{Dooley2024ForecastPFN} proposed the first zero-shot forecasting method trained purely on synthetic data.
% Lag-Llama \citep{Rasul2023LagLlama} introduced a univariate probabilistic forecasting model that was pretrained on a large corpus of diverse time series data.
% TimeFM \citep{Das2024TimesFM} pretrained a decoder style attention model with input patching, using a large time series corpus comprising both real-world and synthetic
% datasets. 
% MOIRAI \citep{Woo2024Unified} is an enhanced Transformer architecture pretrained in the Large-scale Open Time Series Archive, that achieves competitive performance as a zero-shot forecaster.
% \subparagraph{Image-based FMs.} Several works started to explore the use of FMs pretrained on images because of the better intrinsic similarities between images and time series such as trend, stationarity, seasonality/periodicity, and sudden change. 
% \citep{Zhou2023One} tried to fine-tune a BEiT \citep{Bao2021BEiT} trained on images for time series forecasting, but it falls short of the leading text-based and time series-based FMs.
% Recently, VisionTS \citep{Chen2024Visionts} proposes to use a vision Transformer pretrained on ImageNet to reduce the cross-domain gap or in-domain heterogeneity between time series and text.

\paragraph{Vision Models for Time Series.} Several works started to explore the use of FMs pretrained on images because of the better intrinsic similarities between images and time series such as trend, stationarity, seasonality/periodicity, and sudden change. 
\citet{Zhou2023One} tried to fine-tune a BEiT \citep{Bao2021BEiT} trained on images for time series forecasting, but it falls short of the leading text-based and time series-based FMs.
Recently, VisionTS \citep{Chen2024Visionts} proposes to use a vision Transformer pretrained on ImageNet to reduce the cross-domain gap or in-domain heterogeneity between time series and text.

\paragraph{Parameter Efficient Fine Tuning (PEFT).} 
PEFT \citep{peft} is a paradigm to adapt pretrained LLMs to various domains without fine-tuning all of a model’s parameters, which can be costly and require large amounts of training data. % The reader may refer to \cite{PEFTreview} for a comprehensive survey.
% Common PEFT methods include Low Rank Adapters (LoRA) and prompt-based methods. 
LoRA \citep{LoRA} methods freeze the pretrained model weights and injects trainable rank decomposition matrices into each layer of the Transformer architecture, greatly reducing the number of trainable parameters for downstream tasks. 
QLoRA \citep{QLoRA} advances finetuning by significantly reducing memory usage while preserving task performance.

\paragraph{Soft Prompt Tuning.}
Soft prompts have emerged as a compute efficient method for adapting a pretrained LLMs to new domains without altering their core architectures.
\citet{Brown2020Language} were among the first to demonstrate the power of prompting for task adaption of pretrained language models, but automatically finding suitable sets of
text prompts remains an open challenge.
\citet{Li2021Prefix, qin-eisner-2021-learning} proposed the prefix tuning technique that preprends a few task specific soft tokens to the input and hidden states of each Transformer layer. During training, the parameters of soft prompts are updated by gradient descent while the model parameters keep frozen.
\citet{Liu2021Ptuning} showed the prefix tuning technique could be effectively applied to
natural language understanding with different scales of models.
\citet{Lester2021Power} simplified the prefix tuning technique such that it only adds soft prompts to the input layer and is now considered the standard soft prompt-tuning.
% where they optimize sequences of continuous-valued embeddings prepended to the real embeddings of the input tokens.

%% file: section/method.tex
\section{Background} 
% \begin{comment}

\paragraph{Notations.} 
% We first introduce some notations and the problem definition for the tasks that are used in this paper. 
%Following the notations in \cite{Mei2022Transformer},
We observe $n$ events over a fixed time interval $[0, T)$, with each event being denoted as $(e, t)$, where $e \in \mathcal{E}$ is the event type (or attributes) and $\mathcal{E}$ represents the space of event types. An \textit{asynchronous time series} is a sequence of events $x_{1:n} = ((e_1,t_1), (e_2, t_2), \ldots, (e_n, t_n))$ where $t_i$ is an increasing sequence in $[0, T)$ that does not necessarily observe any periodicity. A common alternative to the event time $t_i$ is the inter-arrival time $\tau_j:= t_j - t_{j-1}$; they are considered isomorphic and often used interchangeably. In our work there is very little constraint on $\mathcal{E}$ and in principle, our model still works even if $\mathcal{E}$ is infinite. We only need to be able to compute a vectorial representation of the event type/attributes, which is achieved through the LLM's learned input embeddings in our work.
\vspace{-0.1in}
\paragraph{Language modeling.} 
Language modeling is a widely used task to train LLMs where the goal is predicting the next word or character in a document. Language models are designed to work on a sequence of $m$ tokens, where each token belongs to a vocabulary. 
%$U=(u_1, u_2, \ldots, u_m)$, where $u_i$ is the $i$-th token and belongs to a vocabulary $\mathcal{V}$.
A tokenizer transforms the input text data into a sequence of tokens. The tokenization process is important and can impact performance significantly, for it directly influences how patterns form within tokenized sequences and the types of operations that language models can learn.

%Language models use various statistical and probabilistic techniques to determine the probability of the next token in the sequence, $p_\theta(U) = \prod_{i=1}^m p_\theta(u_i | u_{1:i-1})$.

%\end{comment}

%\todo{Indicate somewhere the tokenizer used for the experiments} \shubham{The tokenizer we use is teh one that comes with Llama3B- Instruct model already, does it need a mention? }

% trained on a collection of sequences, where each sequence is and each token, $u_i$ belongs to a vocabulary. 
\vspace{-0.1in}
\paragraph{Tasks} We propose a new approach to model asynchronous time series with LLMs, which solves three different tasks (see \autoref{fig:overall}).
% \begin{itemize}
% \itemsep0em 
% \item \textbf{Forecasting} (also known as \textit{next event prediction}) Given a history of events $x_{1:m}$ from an asynchronous time series, the model is tasked with predicting the next event $x_{m+1}$.
% %$p(x_{m+1:m+p}|x_{1:m})$. 
% %This definition covers both single even forecasting (when $p=1$) and multi-events forecasting ($p > 1$).
% %The goal of the forecasting task is to predict the future events of the series.
% \item \textbf{Data imputation.} One of the events $x_j$ of the series is randomly chosen and masked, the model is tasked with filling in the gap.
% \item \textbf{Anomaly detection.} One event $x_j$ of the series is randomly chosen and its event type $e_j$ is replaced randomly by another event type $e'$. The model is tasked with identifying this out-of-place element.
% \end{itemize}
\textbf{Forecasting} (also known as \textit{next event prediction}): Given a history of events $x_{1:m}$ from an asynchronous time series, the model is tasked with predicting the next event $x_{m+1}$.
%$p(x_{m+1:m+p}|x_{1:m})$. 
%This definition covers both single even forecasting (when $p=1$) and multi-events forecasting ($p > 1$).
%The goal of the forecasting task is to predict the future events of the series.
\textbf{Data imputation:} One of the events $x_j$ of the series is randomly chosen and masked, and the model is tasked with filling in the gap.
\textbf{Anomaly detection:} One event $x_j$ of the series is randomly chosen and its event type $e_j$ is replaced randomly by another event type $e'$. The model must identify the out-of-place element without knowing the position of the replaced element.

To find the right recipe for the model to solve these tasks, we innovated in two major directions: first, we  studied various representations of the asynchronous time series as inputs to LLMs (Section \ref{ATSrepresentation}) for zero shot completion of these tasks; and secondly, we study different parameter efficient techniques to adapt an LLM backbone for working with asynchronous time series, while leveraging its knowledge of the world and its understanding of natural language (Section \ref{sec:peft_with_lasts}). %We call this method \textit{Stochastic Soft Prompt (StoP)}.

\section{Proposed Method}
% In this paper, we propose 
% \begin{itemize}
%     \item LASTS: A representation of asynchronous time series for LLMs. 
%     \item Parameter-efficient LLM adaptation: Using LoRA, soft prompts, and our novel stochastic soft prompts (StoP), which, in combination with the LASTS representation, achieve state-of-the-art performance across several downstream tasks. \end{itemize}

\subsection{\textbf{LASTS} - Prompting LLMs with Asynchronous Time Series data} \label{ATSrepresentation} 

\begin{comment}
One of our key contributions is the exploration of different methods to represent asynchronous time series data for various tasks.    
\end{comment} 
Unlike ordinary time series, often represented as sequences of numerical values \citep{Gruver2023Large},  asynchronous time series are represented as sequences of events $x_i = (e_i, t_i)$, where $e_i$ is the event type, and $t_i$ is a representation of the timestamp of this event.  Normally, $t_i$ is expressed as an inter-arrival time, which is the time elapsed between event $x_{i-1}$ and $x_i$.

% We will demonstrate that labels in natural language, as opposed to a fixed set of action classes in existing methods, enable the LLM to better understand the context of the problem, leading to better predictions. They also instill a new level of flexibility into the prediction problem, as the input actions are no longer restricted to a particular subset.

% \paragraph{Event representation.}
% We also explore different ways to add dataset description to the prompt.\lilian{TODO: please elaborate}

% \paragraph{Dataset description.}
% \lilian{TODO: please elaborate}
% Understand what is the right recipe/configuration. 

In prior work on modeling asynchronous time series \citep{duTPP, Mehrasa2019Variational, Zhang2020Self, Mei2022Transformer}, events are typically reduced to categories from a small set of options. In contrast, we retain the event types $e_i$ as natural language descriptions. We introduce LASTS, which specifies how to input an asynchronous time series as part of a prompt to effectively leverage LLMs for various tasks on such data.

\begin{figure}[t]
\centering
\includegraphics[width=0.99\columnwidth]{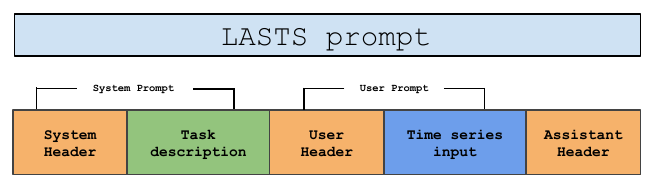}
\caption{Components of a LASTS prompt: A concise task description is included in the system prompt, while asynchronous time series is provided as an input in the user prompt.}
\label{fig:last_prompt}
\end{figure}

\paragraph{LASTS Prompt Structure} The LASTS prompt consists of three parts that can be mapped to the system-user-assistant structure when using an instruction fine-tuned LLM (see \autoref{fig:last_prompt}). The \textbf{system prompt} introduces what an asynchronous time series is, provides a description of the task to be performed, and includes details about the underlying dataset. The \textbf{user prompt} represents the input series as a comma-separated sequence of tuples \((e_i, t_i)\), where \(e_i\) is the textual description of the event type and \(t_i\) is the inter-arrival time. The \textbf{assistant prompt} contains the correct event if performing LLM adaptation training, or is left to be generated by the LLM during inference. More details about the exact prompts used in our experiments can be found in Appendix \ref{sec:zero_shot_prompts}.

% \begin{itemize}

%     \item The \textbf{system prompt} introduces what an asynchronous time series is, provides a description of the task to be performed, and includes details about the underlying dataset;
%     \item The \textbf{user prompt} represents the input series as a comma-separated sequence of tuples \((e_i, t_i)\), where \(e_i\) is the textual description of the event type and \(t_i\) is the inter-arrival time;
%     \item The \textbf{assistant prompt} contains the correct event if performing LLM adaptation training, or is left to be generated by the LLM during inference.
% \end{itemize}
\begin{figure*}[t]
\centering
\includegraphics[width=0.99\textwidth]{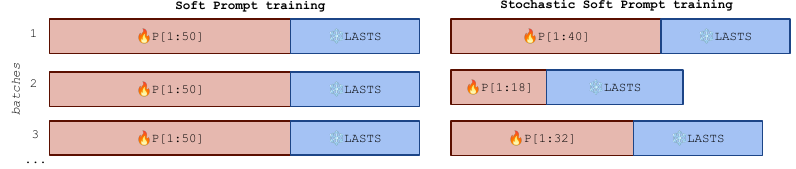}
\caption{Comparison of Soft Prompt (SP) and Stochastic Soft Prompt (StoP) training. For illustration, the soft prompt $P$ is of length $50$. In SP, the entire prompt is used during both training and inference. In StoP, a random prefix of $P$ is used per training batch, while the full prompt is used for inference. Fire marks the soft prompt, which is the trainable prompt portion, while snowflake represents the frozen LASTS text prompt.}
\label{fig:ssp_vs_sp}
\end{figure*}

% \paragraph{Use of Prior World Knowledge by LLMs}

\subsection{Parameter Efficient LLM Adaptation with LASTS representation}
\label{sec:peft_with_lasts}

% \begin{figure*}[t]
% \centering
% \includegraphics[width=0.99\textwidth]{figures/ICLR_Model_2.pdf}
% \caption{Overview of LASTS and Prompt Tuning with LASTS: (a) Structure of a LASTS prompt for adapting an asynchronous time series for use as input for an LLM. (b) $SP$ and $StoP$ training setup - the LLM backbone is frozen and the soft prompt is fine-tuned via gradients computed through the standard next token prediction loss. (c) Comparison of SP and StoP training. In StoP, a random prefix of the trainable prompt is used during each training batch.
% }
% \label{fig:ssp}
% \end{figure*} 

Having established a representation of asynchronous time series for use with LLMs via LASTS, we further enhance the model's adaptability to various tasks using three different adaptation techniques: 

\paragraph{Low Rank Adaption} LoRA is a family of low-rank adaptation techniques that reduce the number of trainable parameters by learning low-rank updates to selected model weights, allowing for efficient fine-tuning of large models. We adapt the LLM backbone for our tasks by applying low-rank adaptations using the LASTS representation as inputs to encode both the task and the input asynchronous time series.

\paragraph{Soft Prompting (SP)} SP involves prepending a continuous prompt to the LASTS representation, which is trained through gradients from next token prediction loss. This guides the model towards task-specific behavior without altering the model weights directly. (See \autoref{fig:ssp_vs_sp})

\paragraph{Stochastic Soft Prompting (StoP)} We propose Stochastic Soft Prompts - an enhancement of SP which learns more robust prompts by imposing a coarse-to-fine structure on the prompt tokens. (See \autoref{sp_vs_stop}).  Similar to SP, we prepend a continuous prompt to the LASTS representation which is trained through gradients from a next-token prediction loss. However, in SP, the entire soft prompt $P$
of length  $L$ is used during training, while in StoP, we randomly select a prefix of the prompt $P$ for each training batch. Specifically, for each batch, we sample a prefix length $l$ from a probability distribution $p(l)$, where $l \leq L$. The soft prompt used for that batch is then represented by $P_{batch} = P[:l] \ \ \text{with} \ \ l \sim p(l)$. In our experiments, we use a uniform distribution as $p$. Both the forward pass and the backward pass are conducted using only the selected prefix $P_{\text{batch}}$. During inference, we use the entire learned soft prompt of length $L$:  $ P_{\text{inference}} = P[1:L] $
See Figure \ref{fig:ssp_vs_sp} for more details. {Our approach is inspired by techniques like dropout \citep{Srivastava2014Dropout} and stochastic depth \citep{huang2016deep}, as well as audio models like SoundStream \citep{Zeghidour2021SoundStream}, where randomly selecting the first $k$ codebooks during training enables better generalization. Similarly, we draw inspiration from Matryoshka Representations \citep{kusupati2022matryoshka}, which learn representations such that predefined prefix lengths remain valid representations.}

These adaptation techniques enable an LLM backbone to handle a variety of asynchronous time series tasks, including forecasting, imputation, and anomaly detection, while maintaining parameter efficiency. Details on the exact prompt representation are provided in Appendix \ref{sec:lasts_for_peft}.

%% file: section/exp.tex
\subsection{Experimental setup}
\label{sec:datasets}
\paragraph{Datasets.} We perform experiments on two different sets of datasets: three text-based action datasets and five standard temporal point process datasets.
The main difference is that actions are represented by words in the action datasets, whereas they are represented by indices in temporal point process datasets.
The \textbf{text-based action datasets} are built from the action annotations of activity videos.  
\textit{Breakfast} \citep{Kuehne2014Language} contains 1712 videos with 177 action classes related to breakfast preparation. 
Each video has a sequence of events to prepare breakfast, with each event containing the timestamp and the action.
\textit{EPIC-KITCHENS-100} \citep{Damen2022Rescaling} is a large-scale dataset in egocentric vision capturing daily activities in the kitchen over multiple days with a total of 100 hours of recording.
It presents more complex activity than the Breakfast dataset, with rich annotations of sequences of actions comprising 97 verb classes and 300 noun classes, with 20k unique narrations. 
\textit{MultiTHUMOS} \citep{Yeung2018Every} contains 400 videos with 65 action classes related to  human activities.
Each video has a sequence of human activity events, with each event containing the timestamp and the activity.
For the \textbf{temporal point process datasets}, we use the five benchmarks introduced in \citep{Xue2023EasyTPP}: 
\textit{Amazon} \citep{Ni2019Justifying} where the goal is to predict the timestamp and category (among 16 categories) of the next reviewed product, 
\textit{Retweet} \citep{Zhou2013Learning} where the goal is to predict the timestamp and category (among 3 categories) of the next user to retweet a post, 
\textit{Taxi} \citep{TaxiData} where the goal is to predict the timestamp and category (among 10 categories) of the next pick-up or drop-off of a taxi driver, 
\textit{Taobao} \citep{Xue2022Hypro} where the goal is to predict the timestamp and category (among 20 categories) of the item clicked by a user, 
and \textit{StackOverflow}\footnote{\scriptsize\url{https://snap.stanford.edu/data/}} where the goal is to predict the timestamp and category (among 22 categories) of the next badges assigned to a given user.
We follow the same data preprocessing as in \citep{Xue2023EasyTPP}.
For each of these datasets, the semantic meaning of the event type is unknown, and only the index of the event type is available.
We use the index of the event type as input to our model.

\paragraph{Metrics.} 
Due to the bi-modal nature of the asynchronous time series, we report separate metrics for the event type and time.
We report the Macro-F1 (M-F1) \citep{Yang1999Evaluation} for event type prediction as Macro-F1 is better suited for multi-class classification tasks with skewed class distributions (Appendix \ref{app:class_imbalance}) than accuracy because Macro-F1 gives equal importance to all the classes. We also report accuracy numbers in Appendix \ref{sec:accuracy_numbers}.
% As shown in \autoref{fig:class_imbalance}, most of the datasets used in this paper exhibit class imbalances, which makes accuracy a less reliable metric. 
% Macro-F1 treats all classes equally by computing the average F1 score across all classes, ensuring that no class is given undue weight, regardless of its frequency in the dataset. This approach provides a more balanced evaluation of model performance across all event types.
We report either the Mean Absolute Error (MAE) or Root Mean Square Error (RMSE) for time prediction.

% \paragraph{Metrics.} 
% Our main evaluation metrics are Macro-F1 for event type prediction, and Mean Absolute Error (MAE) or Root Mean Square Error (RMSE) for time prediction. we use Macro-F1 instead of accuracy for event type prediction, as Macro-F1 is well suited for multi-class classification tasks with potentially skewed class distributions. As shown in Figure \ref{fig:dataset_skew}, the datasets exhibit class imbalances, which makes accuracy a less reliable metric. Macro-F1 treats all classes equally by computing the average F1 score across all classes, ensuring that no class is given undue weight, regardless of its frequency in the dataset. This approach provides a more balanced evaluation of model performance across all event types.

\paragraph{Implementation details} We use Llama-3-8B-Instruct \citep{Dubey2024Llama3} as our LLM backbone. For zero-shot experiments, we disable sampling during response generation, ensuring deterministic outputs. For LLM adaptation experiments, we use QLoRA as the low rank adaptation algorithm, Adam as the optimizer, and a constant learning rate of $2e^{-4}$ for QLoRA and $1e^{-4}$ for prompt tuning. Following \cite{Xue2023EasyTPP}, we split our datasets into a train/validation/test ratio of 70/10/20. Both SP and StoP training are conducted for the same number of epochs. We employ early stopping based on the Macro-F1 on the validation set. We report  performance on the test set.

We use a prompt length of $400$ for prompt tuning in both SP and StoP experiments. This value was selected through hyperparameter tuning across all datasets and tasks, striking a balance between model capacity, performance, and the compute resources available to us. Given that Llama-3-8B-Instruct has a hidden dimension of $4096$, this configuration results in approximately $1.6M$ trainable parameters, which corresponds to only $0.02\%$ of the LLM parameters. For QLoRA, we use a rank of $4$, resulting in a comparable number of trainable parameters ($1.7M$).

\subsection{Experiment Results}

\begin{figure}[h!]
    \centering
    \includegraphics[width=\columnwidth]{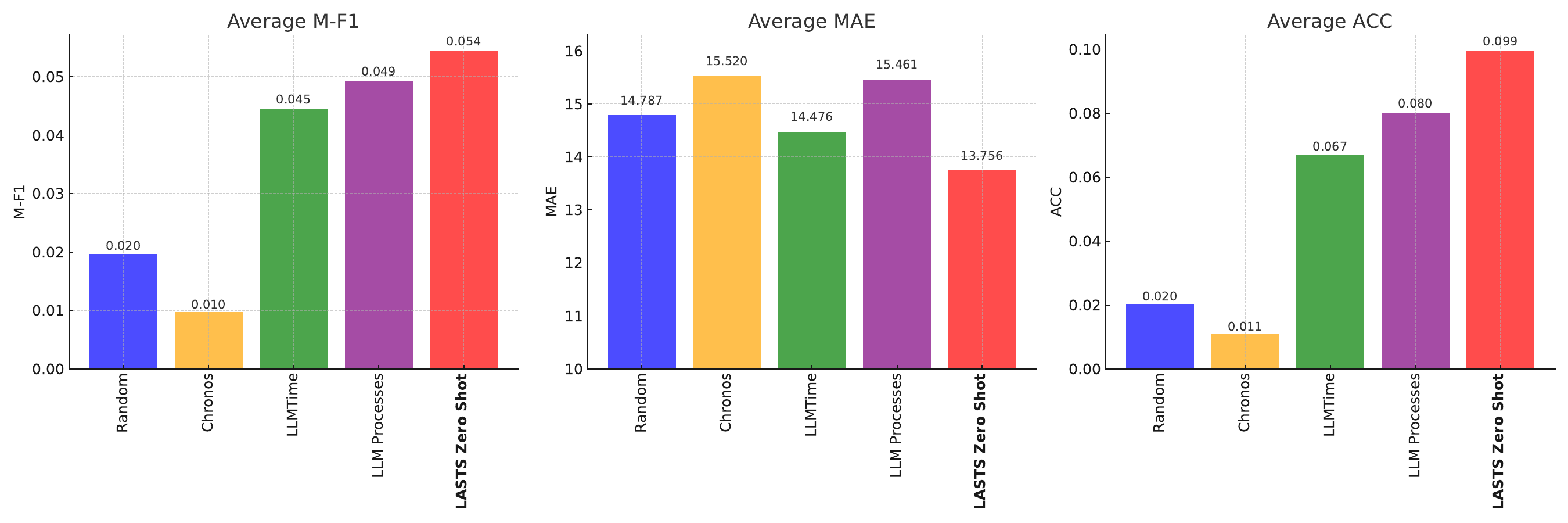}
    \caption{
        Macro-F1 $\uparrow$, MAE $\downarrow$, and Accuracy $\uparrow$, averaged across all datasets for Forecast and Imputation for Zero Shot methods.
    }
    \label{fig:zero_shot_comparison}
\end{figure}

\paragraph{Baselines} We evaluate our methods using four sets of baselines. See Appendix  \ref{sec:llms_for_ts_baselines} for details.
\begin{itemize}
    \item \textbf{Random baseline}: We establish a random baseline simulating random guesses to evaluate our methods on the three text-based datasets and tasks~ (\autoref{tab:performance_evaluation_new}, \autoref{fig:zero_shot_comparison}).
    \item \textbf{Foundation models for time series}: We use a state-of-the-art pretrained foundation model for time series forecasting, \textbf{Chronos} \cite{ansari2024chronos}, as a baseline for forecasting and imputation tasks on asynchronous time series~ (\autoref{tab:performance_evaluation_new}).
    \item \textbf{LLM for time series}:  We adapt two LLM-based time series forecasting methods, \textbf{LLMTime} \citep{Gruver2023Large} and \textbf{LLMProcesses} \citep{requeima2024llm}, as baselines for zero-shot LASTS prompting on asynchronous time series~ (\autoref{tab:performance_evaluation_new}, \autoref{fig:zero_shot_comparison}).
    \item \textbf{TPP models}: We compare our model with state-of-the-art TPP models for asynchronous time series \citep{Xue2023EasyTPP}. We report the results for two popular RNN-based models: Recurrent marked temporal point process (RMTPP) \citep{duTPP} and neural Hawkes Process (NHP) \citep{MeiEisner}. We also compare with three attention-based models: self-attentive Hawkes process (SAHP) \citep{Zhang2020Self}, Transformer Hawkes process (THP) \citep{Zuo2020Transformer}, attentive neural Hawkes process (AttNHP) \citep{Yang2022Large}~ (\autoref{tab:main_results_pred_main_text}).
\end{itemize}
\vspace{-0.1in}
\paragraph{Results} Our results on the the three tasks (forecast, imputation, anomaly detection) and the three  text datasets (Breakfast, MultiTHUMOS, EPIC-KITCHENS) are presented in \autoref{tab:performance_evaluation_new}. Based on our results, we make five main observations. Firstly, {LASTS proves to be an effective and robust representation for asynchronous time series data across multiple datasets. LASTS Zero Shot consistently outperforms the Time Series Foundation Model Chronos and LLM-based methods (LLMTime and LLM Processes) in most evaluations, highlighting the advantage of using textual event descriptions enabled by LASTS.}  Secondly, our results demonstrate that the LASTS representation can be applied across multiple tasks without any investment needed in designing custom models for each task. Thirdly, LASTS work effectively with multiple LLM adaptation techniques without algorithm-specific modifications. Fourthly, we observe that StoP as an adaptation technique outperforms other techniques for most time prediction evaluations, and in all event type prediction evaluations. Finally, we highlight our results on the EPIC-KITCHENS dataset, which features very rich textual event descriptions (approximately 20,000). While traditional TPP modeling methods struggle to handle such a large set of classes, our approach effectively models various tasks on this complex dataset.

\begin{table}[htb]
    \centering
    % \caption{\textit{Performance evaluation on three textual datasets across forecasting, imputation, and anomaly detection tasks. Metrics are macro F1 (M-F1) and Mean Absolute Error (MAE) where applicable. The \textbf{best result} in each category is highlighted in bold, and the \underline{second-best result} is underlined. Note that the TPP based methods NHP, SAHP and AttNHP are only designed for forecasting. Note that for anomaly detection, since the task involves identifying only the anomalous event, the MAE metric is not applicable and Chronos and LLMProcesses are not adaptable (see  \ref{sec:llms_for_ts_baselines}). A $^*$ indicates our method. We use $5$ examples for few shot results (see \ref{sec:few-shot-analysis}).}}
    \caption{\textit{Performance evaluation on three textual datasets for forecasting, imputation, and anomaly detection. Metrics: Macro F1 (M-F1) and Mean Absolute Error (MAE) where applicable. \textbf{Best} results are in bold, \underline{second-best} are underlined. For anomaly detection, MAE is inapplicable, and Chronos/LLMProcesses are non-adaptable (see \ref{sec:llms_for_ts_baselines}). A $^*$ indicates our method. Few-shot results use five examples (see \ref{sec:few-shot-analysis}).}
}
    \resizebox{\columnwidth}{!}{
    \begin{tabular}{l|cc|cc|cc}
        \toprule
        \multicolumn{7}{c}{\textbf{Forecasting}} \\
        \midrule
        \textbf{Model} 
        & \multicolumn{2}{c|}{\textbf{Breakfast}} 
        & \multicolumn{2}{c|}{\textbf{MultiTHUMOS}} 
        & \multicolumn{2}{c}{\textbf{EPIC-KITCHENS}} \\
        & M-F1 $\uparrow$ & MAE $\downarrow$ 
        & M-F1 $\uparrow$ & MAE $\downarrow$ 
        & M-F1 $\uparrow$ & MAE $\downarrow$ \\
        \midrule
        Random & 0.0162 & 40.1513 & 0.0417 & 1.8803 & 0.0000 & 3.2001 \\
        Chronos & 0.0011 & 43.0502 & 0.0265 & 1.9805 & 0.0000 & 3.5925 \\
        LLMTime & 0.0240 & 37.3902 & 0.1280 & 2.2060 & 0.0040 & 4.8948 \\
        LLMProcesses & 0.0337 & 44.9856 & 0.1278 & 2.0471 & 0.0049 & 4.3843 \\
        LASTS Zero Shot$^*$ & 0.0604 & 38.1630 & 0.1361 & 1.8868 & 0.0105 & 3.1566 \\
        LASTS Few Shot$^*$ & 0.1518 & 35.5605 & 0.1676 & 1.8114 & 0.0149 & 3.3092 \\
        LASTS + QLORA$^*$ & \underline{0.2558} & 33.9737 & 0.3218 & 1.7281 & 0.0764 & \underline{2.8964} \\
        LASTS + SP$^*$ & 0.2341 & \underline{32.8417} & \underline{0.3707} & \underline{1.6630} & \underline{0.0780} & \textbf{2.8830} \\
        LASTS + StoP$^*$ & \textbf{0.2633} & \textbf{32.5464} & \textbf{0.3947} & \textbf{1.6503} & \textbf{0.0797} & 3.0318 \\
        \midrule
        \multicolumn{7}{c}{\textbf{Imputation}} \\
        \midrule
        \textbf{Model} 
        & \multicolumn{2}{c|}{\textbf{Breakfast}} 
        & \multicolumn{2}{c|}{\textbf{MultiTHUMOS}} 
        & \multicolumn{2}{c}{\textbf{EPIC-KITCHENS}} \\
        & M-F1 $\uparrow$ & MAE $\downarrow$ 
        & M-F1 $\uparrow$ & MAE $\downarrow$ 
        & M-F1 $\uparrow$ & MAE $\downarrow$ \\
        \midrule
        Random & 0.0168 & 37.7029 & 0.0435 & 2.3622 & 0.0000 & 3.4269 \\
        Chronos & 0.0013 & 38.4039 & 0.0294 & 2.3971 & 0.0000 & 3.6955 \\
        LLMTime & 0.0137 & 35.9899 & 0.0968 & 2.6998 & 0.0005 & 3.6750 \\
        LLMProcesses & 0.0156 & 34.7117 & 0.1123 & 2.3786 & 0.0008 & 4.2600 \\
        LASTS Zero Shot$^*$ & 0.0263 & 33.0097 & 0.0915 & 2.6696 & 0.0015 & 3.6527 \\
        LASTS Few Shot$^*$ & 0.0520 & 33.3440 & 0.1013 & 2.3982 & 0.0023 & 3.2528 \\
        LASTS + QLORA$^*$ & 0.1688 & \underline{28.5638} & \underline{0.2132} & \textbf{2.2179} & 0.0378 & \underline{3.1194} \\
        LASTS + SP$^*$ & \underline{0.1581} & 28.8503 & 0.2044 & 2.4092 & \underline{0.0423} & 3.1456 \\
        LASTS + StoP$^*$ & \textbf{0.2064} & \textbf{28.2251} & \textbf{0.2213} & \underline{2.3445} &  \textbf{0.0610} & \textbf{3.1116} \\
        \midrule
        \multicolumn{7}{c}{\textbf{Anomaly Detection}} \\
        \midrule
        \textbf{Model} 
        & \multicolumn{2}{c|}{\textbf{Breakfast}} 
        & \multicolumn{2}{c|}{\textbf{MultiTHUMOS}} 
        & \multicolumn{2}{c}{\textbf{EPIC-KITCHENS}} \\
        & M-F1 $\uparrow$ & MAE $\downarrow$ 
        & M-F1 $\uparrow$ & MAE $\downarrow$ 
        & M-F1 $\uparrow$ & MAE $\downarrow$ \\
        \midrule
        Random & 0.0349 & \textemdash & 0.0381 & \textemdash & 0.0238 &  \textemdash \\
        LLMTime & 0.0240 &  \textemdash & 0.0415 &  \textemdash & 0.0048 &  \textemdash\\
        LASTS Zero Shot$^*$ & 0.0923 & \textemdash & 0.2755 &  \textemdash & 0.0159 &  \textemdash \\
        LASTS Few Shot$^*$ & 0.0837 &  \textemdash  & 0.3535 &  \textemdash & 0.0337 &  \textemdash \\
        LASTS + QLORA$^*$ & \underline{0.7011} &  \textemdash & \underline{0.6003} &  \textemdash & \underline{0.6520} &  \textemdash \\
        LASTS + SP$^*$ & 0.6520 &  \textemdash & 0.5231 &  \textemdash & 0.6159 &  \textemdash \\
        LASTS + StoP$^*$ & \textbf{0.7198} &  \textemdash & \textbf{0.6045} &  \textemdash & \textbf{0.6603} &  \textemdash \\
        \bottomrule
    \end{tabular}
    }
    \label{tab:performance_evaluation_new}
\end{table}

\begin{table*}[h]
\caption{\textit{Performance of models on next-event's type and type prediction across five real datasets.  The \textbf{best result} is shown in bold, and the \underline{second best result} is underlined. OOM indicates an Out Of Memory error. A missing entry indicates the model diverged. We tried optimizing these baselines for the three textual datasets—MultiTHUMOS (65 classes), Breakfast (177 classes), and EPIC-KITCHENS ($\sim$ 20K classes)—but these models either diverged, performed poorly, or ran out of memory due to the large number of classes.}}
\centering
\resizebox{\textwidth}{!}{
\begin{tabular}{|l|cc|cc|cc|cc|cc|cc|cc|cc|}
\toprule
Model & \multicolumn{2}{c|}{Amazon} & \multicolumn{2}{c|}{Retweet} & \multicolumn{2}{c|}{Taxi} & \multicolumn{2}{c|}{Taobao} & \multicolumn{2}{c|}{StackOverflow} & \multicolumn{2}{c|}{Breakfast} & \multicolumn{2}{c|}{MultiTHUMOS} & \multicolumn{2}{c|}{EPIC-KITCHENS}\\
\cmidrule(lr){2-17}
& M-F1 $\uparrow$ & RMSE $\downarrow$ & M-F1 $\uparrow$ & RMSE $\downarrow$ & M-F1 $\uparrow$ & RMSE $\downarrow$ & M-F1 $\uparrow$ & RMSE $\downarrow$ & M-F1 $\uparrow$ & RMSE $\downarrow$ & M-F1 $\uparrow$ & RMSE $\downarrow$ & M-F1 $\uparrow$ & RMSE $\downarrow$ & M-F1 $\uparrow$ & RMSE $\downarrow$\\
\midrule
RMTPP    & 0.0988 & \underline{0.4780} & 0.3110 & 16.5849 & 0.2969 & 0.3761 & \underline{0.4495} & \underline{0.1338} & 0.0277 & 1.3727  & - & - & - & - & OOM & OOM \\	
NHP      & 0.1266 & \textbf{0.4489} & 0.4128 & \textbf{15.6233} & \underline{0.3667} & 0.3995 & 0.4287 & 0.1822 & 0.0559 & 1.3960   & 0.0167 & 116.23 & \underline{0.2861} & 4.8583 & OOM & OOM\\
SAHP     &  0.0846 & 0.5491 & 0.2772 & 16.6451 & 0.2780 & \textbf{0.3193} & 0.1816 & 0.1347 & 0.0322 & \underline{1.3326}   & 0.0023 &  112.85& 0.0 &  \underline{4.5908} & OOM & OOM\\
THP      &  \underline{0.1414} & 0.4911 & 0.2114 & 16.6440 & 0.3451 & 0.3736 & 0.2734 & 0.1340 & \underline{0.0661} & 1.4054   & - & - & - & - & OOM & OOM\\
AttNHP   &  0.1270 & 0.7054 & \underline{0.4210} & 16.8278 & 0.2167 & 0.4072 & 0.1048 & 0.1350 & 0.0475 & 1.3661  & \underline{0.0478} & \underline{108.41} & 0.0809 & 5.2113 & OOM & OOM \\
    \midrule
% \textbf{SP}       &  0.143 & 0.601 & \textbf{0.4331} & 16.4713 & 0.3204 & 0.3314 & \textbf{0.4652} & \textbf{0.1321} & 0.0943 & 1.2657 & & & & & & \\
\textbf{LASTS + StoP}      & \textbf{0.1520} & 0.6000 & \textbf{0.4299} & \underline{16.4981} &\textbf{0.4174} & \underline{0.3278} & \textbf{0.4633} & \textbf{0.1321} & \textbf{0.0983} & \textbf{1.2596} & \textbf{0.2633} & \textbf{102.02} & \textbf{0.3947} & \textbf{3.6722} & \textbf{0.0797} & \textbf{7.3724} \\
\bottomrule
\end{tabular}
}
\label{tab:main_results_pred_main_text}
\end{table*}
\vspace{-0.1in}

\vspace{-0.1in}
\paragraph{Comparison with TPP models.} 
\autoref{tab:main_results_pred_main_text} shows experimental results that compare our model with existing TPP models on standard TPP datasets.
% We report the results for two popular RNN-based models: Recurrent marked temporal point process (RMTPP) \citep{duTPP} and neural Hawkes Process (NHP) \citep{MeiEisner}.
% We also compare with three attention-based models: self-attentive Hawkes process (SAHP) \citep{Zhang2020Self}, Transformer Hawkes process (THP) \citep{Zuo2020Transformer}, attentive neural Hawkes process (AttNHP) \citep{Yang2022Large}.
% To the best our knowledge, our model is the first LLM-based model for asynchronous time series so we cannot compare with other LLM-based models.
TPP models are designed for forecasting so we only show the results for the forecasting task.
% \todo{I assume our model is the best on Breakfast, MH, and EK. We will need to adapt the description if it is not true.} \shubham{It  is true. These methods dont give baseline for EPIC-KITCHENS due to large category space and on MH and BK our method is best.}
We observe that our model has competitive results w.r.t.~TPP models, 
outperforming existing TPP models on 13 of the 18 evaluations, and is in the top-2 best models on 17 of the 18 evaluations. 
Our model has the best performance for all the event type evaluations, which shows that our model is more accurate to predict the next event type. 
On three of the eight datasets, our model is less accurate than TPP models to predict the time.
We think that our model is not performing as well as the TPP models, because our model does not have an explicit prior about the time distribution whereas TPP models (e.g.~Poisson process or Hawkes process) make strong assumptions about the time distribution. {In the case of the Amazon dataset, the performance gap is more pronounced because this dataset groups a large number of diverse event types into a single event category, making it harder to model inter-arrival times.} 
These results show that our model is able to outperform existing TPP models on most of the datasets without explicit modeling of the time distribution.
% We think it may be possible to improve the performance of our model by adding a distribution prior in the prompt, and leave it as future work. 
It also shows that our model is performing well even when only the index of the event type is provided instead of its textual description, making it a more generally applicable method~(See Appendix \ref{sec:llms_for_ts_baselines}).
% \vspace{-0.2in}
\paragraph{Comparison with Zero Shot Methods} \autoref{fig:zero_shot_comparison} shows LASTS Zero Shot outperforms other zero shot techniques over all metrics when averaged over all tasks and datasets. See Appendix \ref{sec:llms_for_ts_baselines} for details.
\vspace{-0.1in}
\paragraph{Comparison with PEFT Techniques.} As detailed in Appendix \ref{app:stop_comparison_with_peft}, Stochastic Soft Prompting provides a significant advantage, achieving an average Macro-F1 improvement of $12.69\%$ over vanilla Soft Prompting and $13.55\%$ over QLoRA across all tasks and datasets.
% We compare our model with existing TPP models on standard TPPs datasets and the results are summarized in \autoref{tab:main_results_pred_main_text}.

\begin{figure}[h]
    \centering
    % First Row: t-SNE Projections
    \begin{minipage}[t]{0.45\columnwidth}
        \centering
        \includegraphics[width=\linewidth]{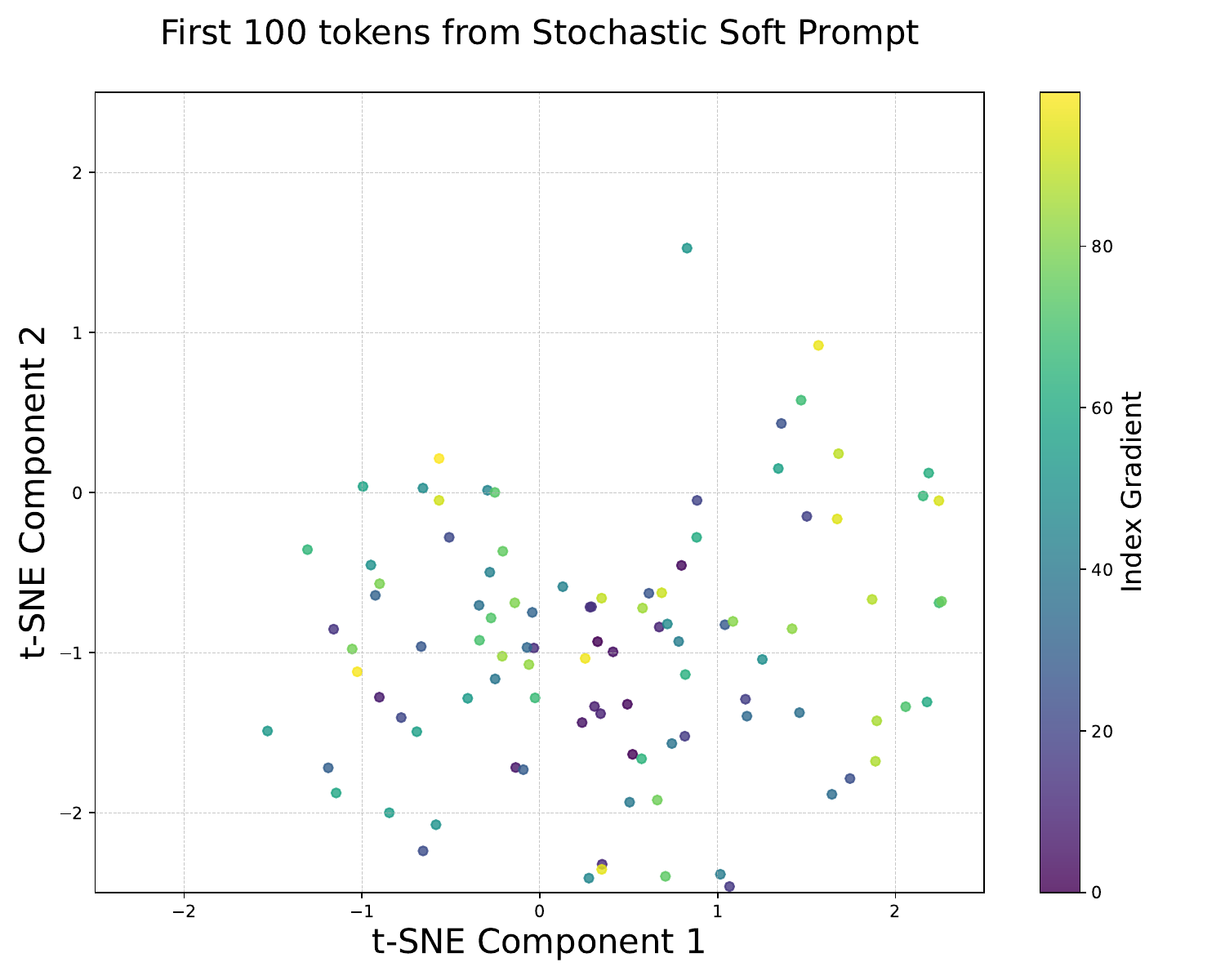}
    \end{minipage}
    \begin{minipage}[t]{0.45\columnwidth}
        \centering
        \includegraphics[width=\linewidth]{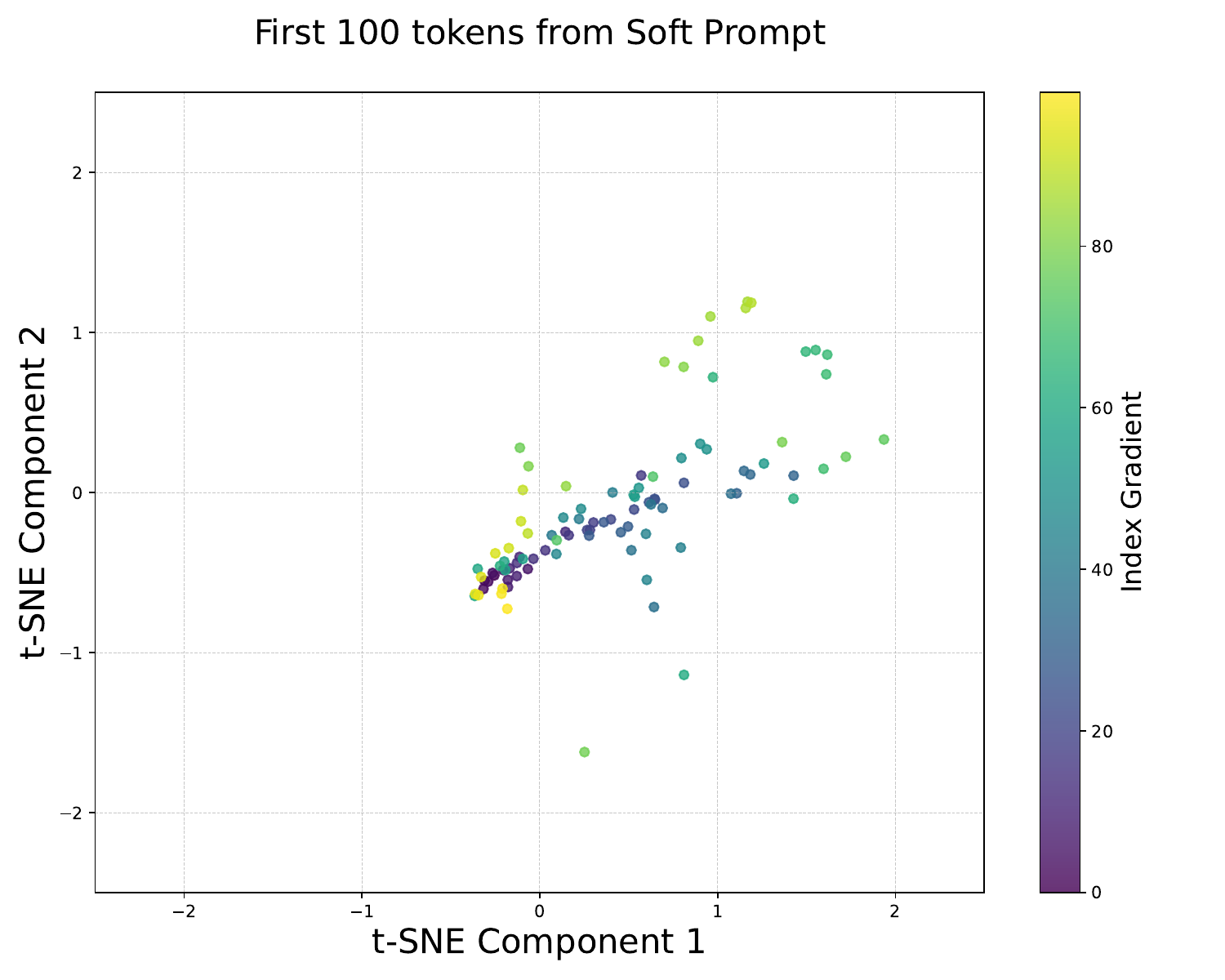}
    \end{minipage}
    \begin{minipage}[t]{\columnwidth}
        \centering
        \includegraphics[width=\linewidth]{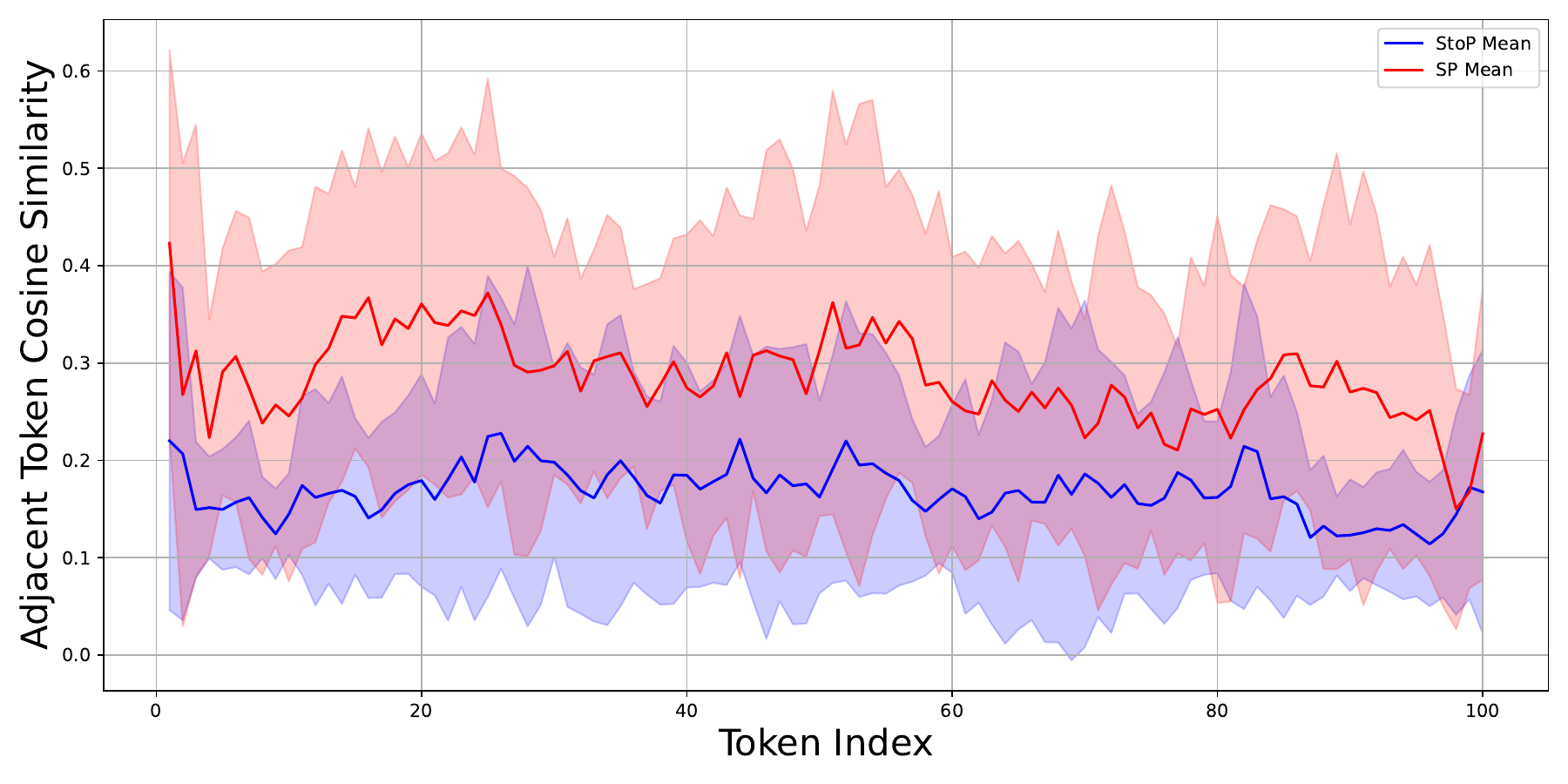}

    \end{minipage}
    
\caption{Learned token representations of StoP and SP. The first two plots show t-SNE projections of 100 tokens from 400-length prompts (Breakfast dataset, forecasting)—StoP tokens are more dispersed, while SP tokens cluster closely. The third plot shows lower adjacent token cosine similarity for StoP (blue) than SP (red), indicating greater diversity.}

\label{fig:tsne}
\end{figure}
\vspace{-0.15in}
\subsection{Model analysis} 
\begin{figure}[t]
    \centering
    \begin{minipage}[t]{0.45\columnwidth}
        \centering
        \includegraphics[width=\linewidth]{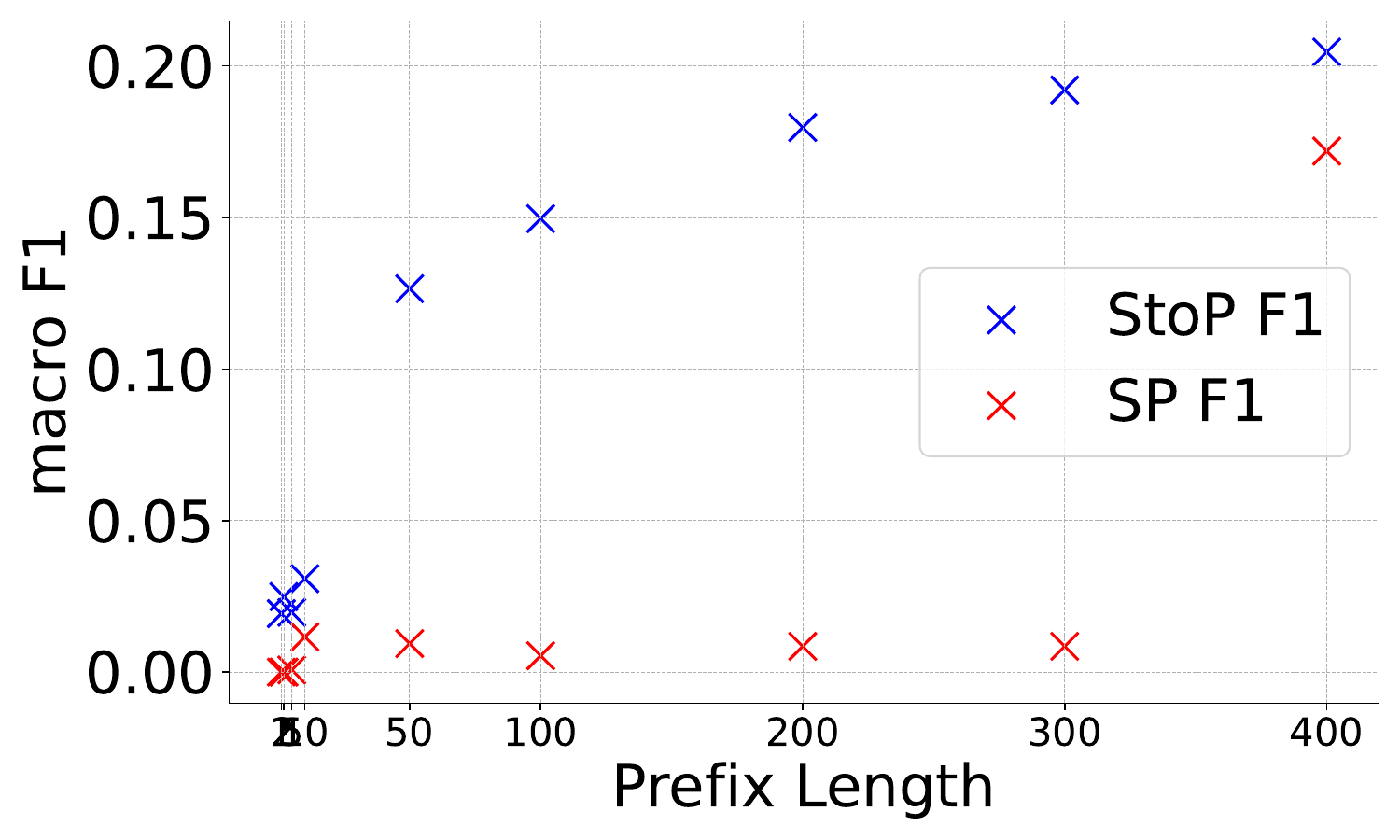}
    \end{minipage}
    \hfill
    % \begin{minipage}[t]{0.31\columnwidth}
    %     \centering
    %     \includegraphics[width=\linewidth]{figures/pdf_plots/acc_comparison.pdf}
    % \end{minipage}\hfill
    \begin{minipage}[t]{0.45\columnwidth}
        \centering
        \includegraphics[width=\linewidth]{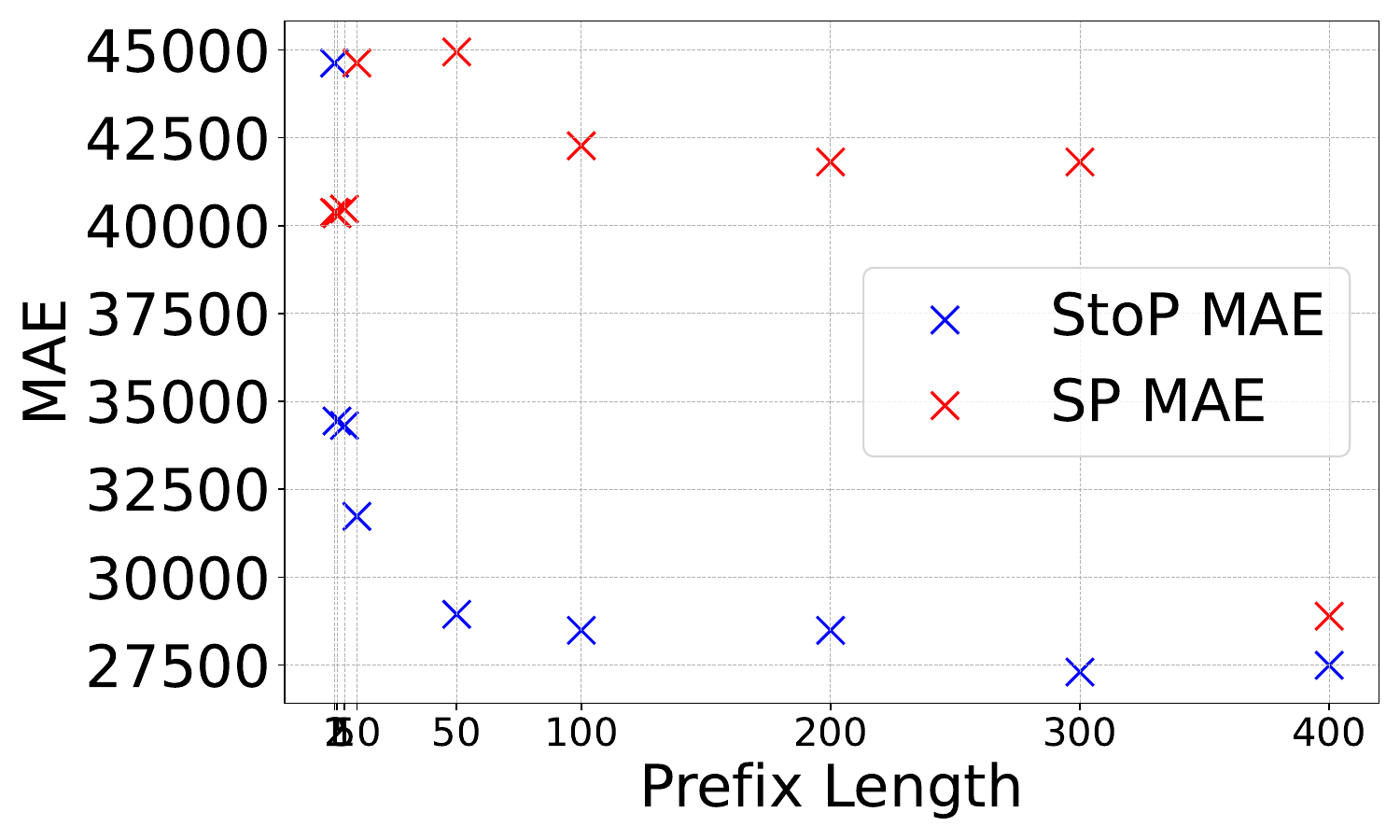}
    \end{minipage}
    
    \caption{StoP-trained prefixes function as standalone prompts, unlike SP. Testing 400-length prompts (Breakfast, imputation) shows StoP prefixes remain effective, while SP prefixes do not.}
    \label{fig:vallid_porefixes}
\end{figure}

% \vspace{-0.05in}
\paragraph{Comparison of SP and StoP learned token representations.} \label{sp_vs_stop}
% \paragraph{Understanding differences in token representations learned by SP and StoP.} 
% The tokens learned by Stochastic Soft Prompt (StoP) and Soft Prompt (SP) have distinct characteristics due to differences in their training paradigms. 
% To illustrate this difference, we plot the t-SNE projections of the first 100 tokens from a prompt of length 400 for both StoP and SP in \autoref{fig:tsne}. 
% We observe that the tokens learned through StoP training are more spread out, indicating greater diversity, while those learned through SP training tend to cluster more closely. 
% StoP uses a coarse-to-fine approach, where the first embeddings are more diverse to cover a large part of the space than the first embeddings trained with SP.
% This difference is further highlighted by the cosine similarity between adjacent tokens in the last plot of \autoref{fig:tsne}: the adjacent tokens in StoP prompts have lower similarity compared to SP. It allows StoP to work better than SP, even when only the first soft tokens are used (see \autoref{fig:vallid_porefixes}). Using more soft tokens further improves StoP, as it gains access to more fine-grained information.
Stochastic Soft Prompt (StoP) and Soft Prompt (SP) learn distinct token distributions due to differences in training. Figure~\ref{fig:tsne} shows t-SNE projections of the first 100 tokens from 400-length prompts. We observe that the tokens learned through StoP training are more spread out, indicating greater diversity, while those learned through SP training tend to cluster more closely. StoP follows a coarse-to-fine approach, with early embeddings that are diverse and cover a larger space. This difference is further highlighted in the last plot of Figure~\ref{fig:tsne}, where StoP tokens have lower adjacent cosine similarity than SP. As a result, StoP outperforms SP even when using only the first few tokens, with further improvements as more tokens are utilized (Figure~\ref{fig:vallid_porefixes}).

% Figure \ref{fig:tsne} shows that adjacent tokens in StoP prompts have lower similarity compared to SP. 
% The StoP training paradigm introduces not only stochasticity but also the selection of a prefix of tokens during training, forcing the tokens to spread apart, localizing learned information to specific positions, possibly making it easier for LLMs to retrieve it.
% \todo{Thibaut, Lilian - does this look ok? Would love animproved language here around whats happening.}
\vspace{-0.05in}
\paragraph{All prefixes are valid prompts in StoP} The training paradigm of StoP forces all prefixes of StoP to act as valid standalone prompts, as they are used as prompts during training for some batches (if trained for long enough). (see \autoref{fig:vallid_porefixes}). This further strengthens our belief that tokens in StoP are arranged from coarse, independent tokens at the beginning to tokens with tokens containing finer information towards the end. See Appendix \ref{sec:stop_analysis} for further discussion.

\vspace{-0.05in}
\paragraph{Disentangling Stochasticity and Prefix Picking in StoP.}  
To highlight the impact of structured prefix picking in StoP, we compare it with an alternative approach where, instead of selecting a prefix, we randomly select $l$ tokens from the prompt per batch, with $l$ drawn from a uniform distribution. We find that stochasticity alone is insufficient for learning effective soft prompts, and structured prefix picking plays a crucial role in StoP’s performance gains~(Appendix \ref{app:stop_disentangling}).

\vspace{-0.07in}
\paragraph{Training speed.} 
Another dimension on which to compare SP and StoP is the training speed. 
Due to differences in training paradigms, StoP trains significantly faster than SP for the same prompt length, as many training batches use only a subset of the full prompt in StoP.
In our experiments with 400 soft prompts, we observed that StoP trains approximately \textbf{$25\%$} faster than SP.
\vspace{-0.07in}
\paragraph{Understanding StoP prompts through probing.} 
% \cite{Lester2021Power} attempts to interpret learned prompts by mapping them to the closest input embeddings, but this often yields incoherent results.  Instead, we treat prompt interpretability as a probing task for the LLM itself, appending the learned prompt with a simple instruction: \textit{"Tell me in as much detail as possible what task you are supposed to do."} This approach allows the LLM to articulate its interpretation directly, providing a more human-understandable explanation of the learned task. For example, this interpretation of a prompt learned on forecasting for breakfast dataset, indicates the high level dataset and task information is learned by the prompt:
While prior work such as \cite{Lester2021Power} attempts to interpret learned prompts by mapping them to the closest input embeddings—often yielding incoherent results—we instead explore probing the LLM using the learned prompt. By appending the learned prompt with a simple instruction, such as \textit{``Tell me in as much detail as possible what task you are supposed to do,''} we encourage the LLM to generate an output that reflects its understanding of the task. This approach allows us to gain some insight into what the model has summarized from the tasks and datasets it has been trained on. We present multiple model responses when probed like this in Appendix \ref{sec:prompt_interpretations_word_embeddings}.
\begin{figure}[h]
    \centering
    \includegraphics[width=0.8\columnwidth]{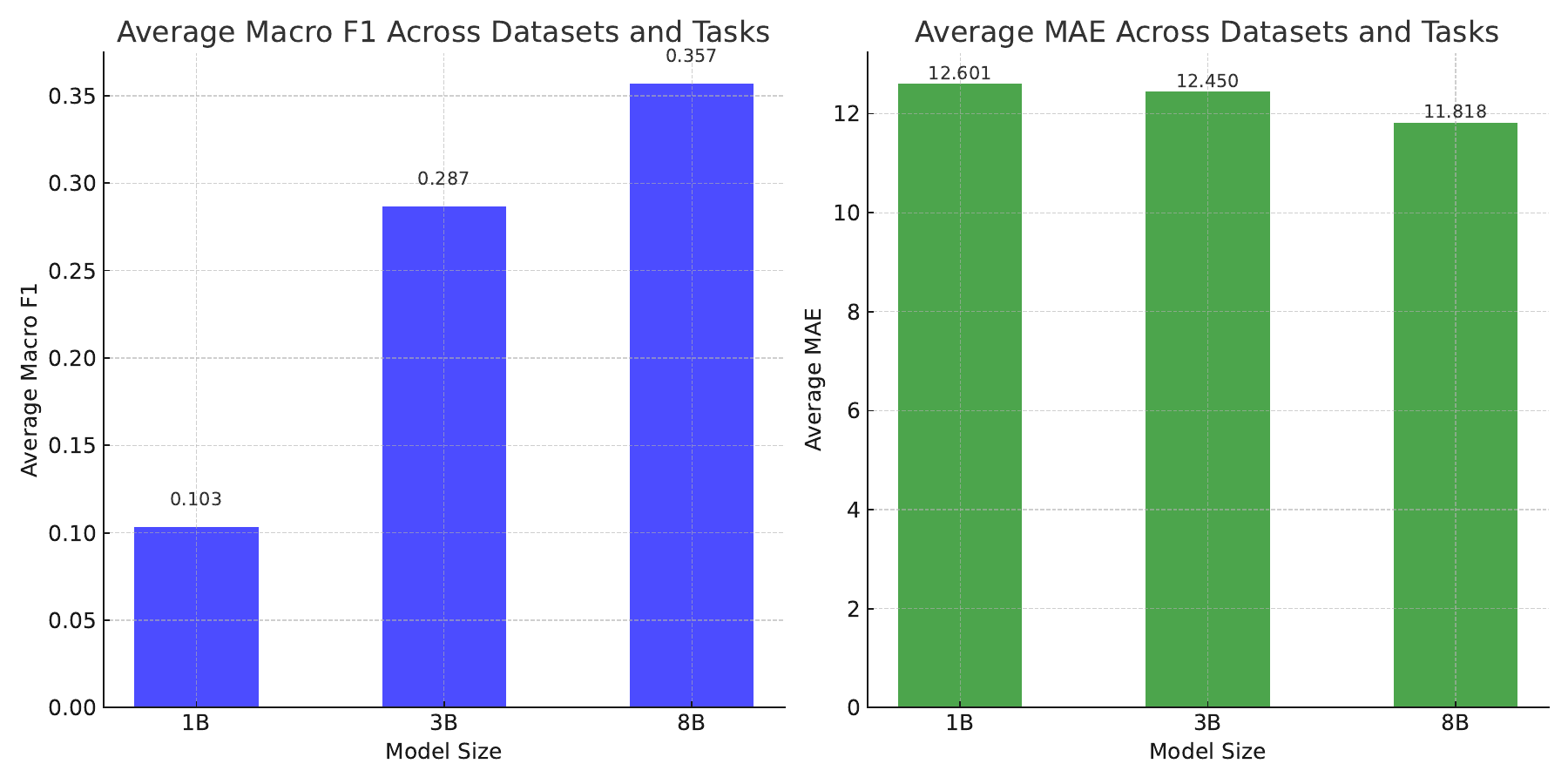}
    \caption{
        Macro-F1 $\uparrow$ and MAE $\downarrow$ across all datasets and tasks for different model sizes.
    }
    \label{fig:avg_macro_f1_mae_by_model_size_paper}
\end{figure}
\vspace{-0.2in}
\paragraph{Scaling Laws.}  
We evaluate Stochastic Soft Prompts (StoP) across different LLM backbone sizes (1B, 3B, and 8B) and observe consistent performance gains with larger models, indicating that StoP benefits from improvements in the underlying LLMs and is expected to scale accordingly~ (\autoref{fig:avg_macro_f1_mae_by_model_size_paper}, Appendix \ref{app:scaling_laws}).

%% file: section/conclusion.tex
\section{Conclusion and Future Work}
\label{sec:conclusion}

We presented a novel approach to modeling asynchronous time series with an LLM, introducing a flexible alternative to  traditional TPP methods. By encoding an asynchronous time series in a prompt, our approach enables LLMs to leverage their world knowledge for various downstream tasks, including forecasting, anomaly detection, and imputation.

Additionally, we proposed Stochastic Soft Prompt (StoP), an efficient PEFT technique for adapting LLMs to asynchronous time series data. This approach not only improves adaptability but also suggests broader applicability to other data modalities such as image or natural language sequences.

Our findings highlight the potential of LLM-based representations for asynchronous time series and suggest new directions for future research, including refining LLM adaptation strategies and exploring hybrid approaches that combine neural architectures with prompt-based modeling.

%% file: section/appendix.tex
\section{Appendix}

\subsection{Dataset Preparation} 
\label{sec:dataset-preparation}

We remove any sequence in the dataset that is very small ($< 4$ elements). We split the dataset in a random $70/10/20$ train, validation and test split. Each sequence is expanded into multiple sequences based on the task:
\begin{itemize}
    \item \textbf{Forecasting}: We convert a sequence into multiple prediction tasks. For each element of the series, the prediction task is to predict the element given the preceding elements. We impose a minimum and maximum length requirements on the number of preceding elements used.
    \item \textbf{Imputation}: For every element in the series, we replace the element by a mask, and the imputation task is to predict the masked element given the remaining sequence.
    \item \textbf{Anomaly Detection}: For every element in the sequence, we replace the action by a random different action. the anomaly detection task is to identify the element of the sequence that has been tampered with.
\end{itemize}

For the three test based datasets - Breakfast, MultiTHUMOS and EPIC-KITCHENS, the event types are already represented as text. The remaining $5$ datasets from the temporal point processes domain lack a textual component, and the event types are represented by integers. For these datasets, we simply treat each integer event type as a string, allowing the LLM to process it similarly to text-based data.

\subsection{Dataset Class Imbalance}
\label{app:class_imbalance}
\begin{figure}[h]
    \centering
    \begin{minipage}[b]{0.45\textwidth}
        \includegraphics[width=\textwidth]{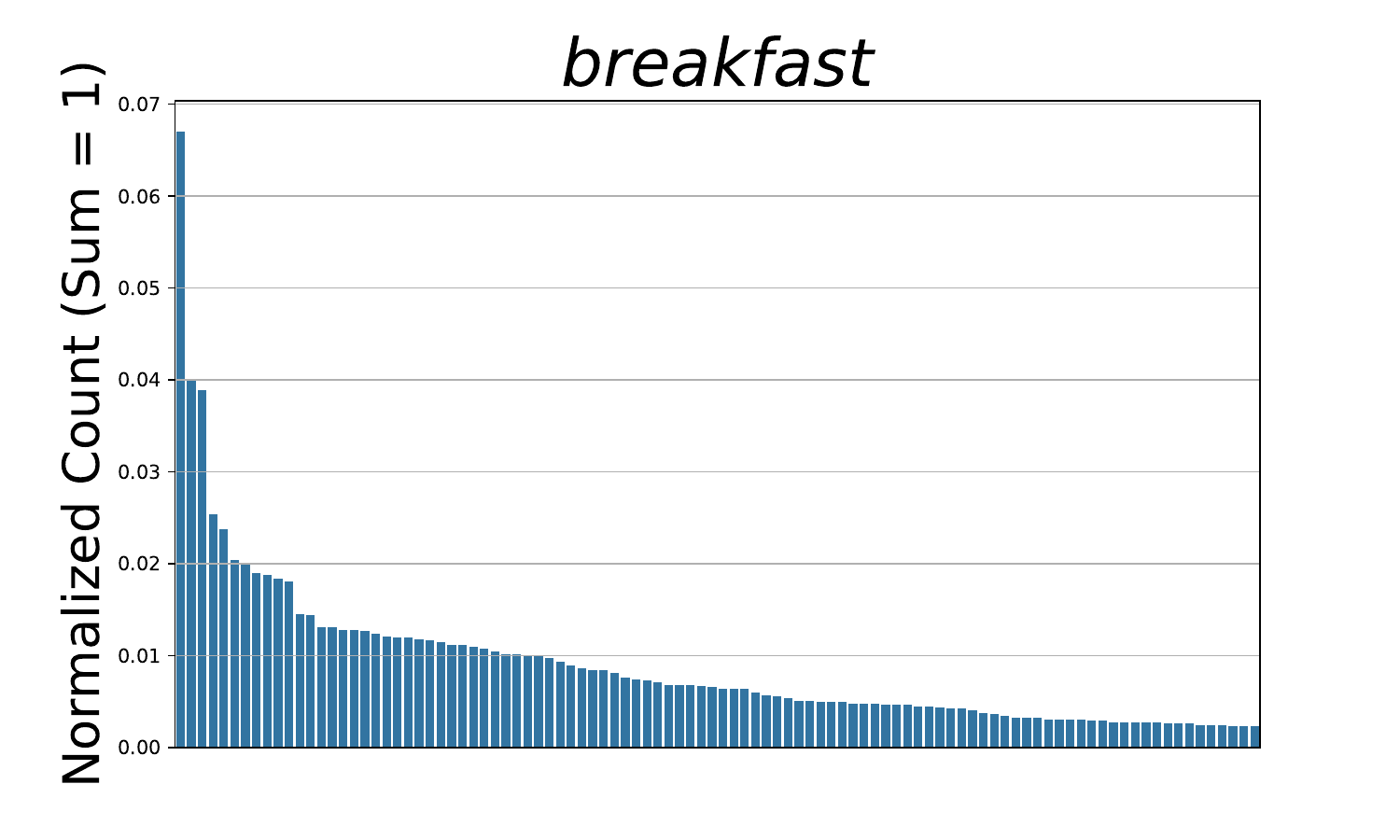}
    \end{minipage}
    \begin{minipage}[b]{0.45\textwidth}
        \includegraphics[width=\textwidth]{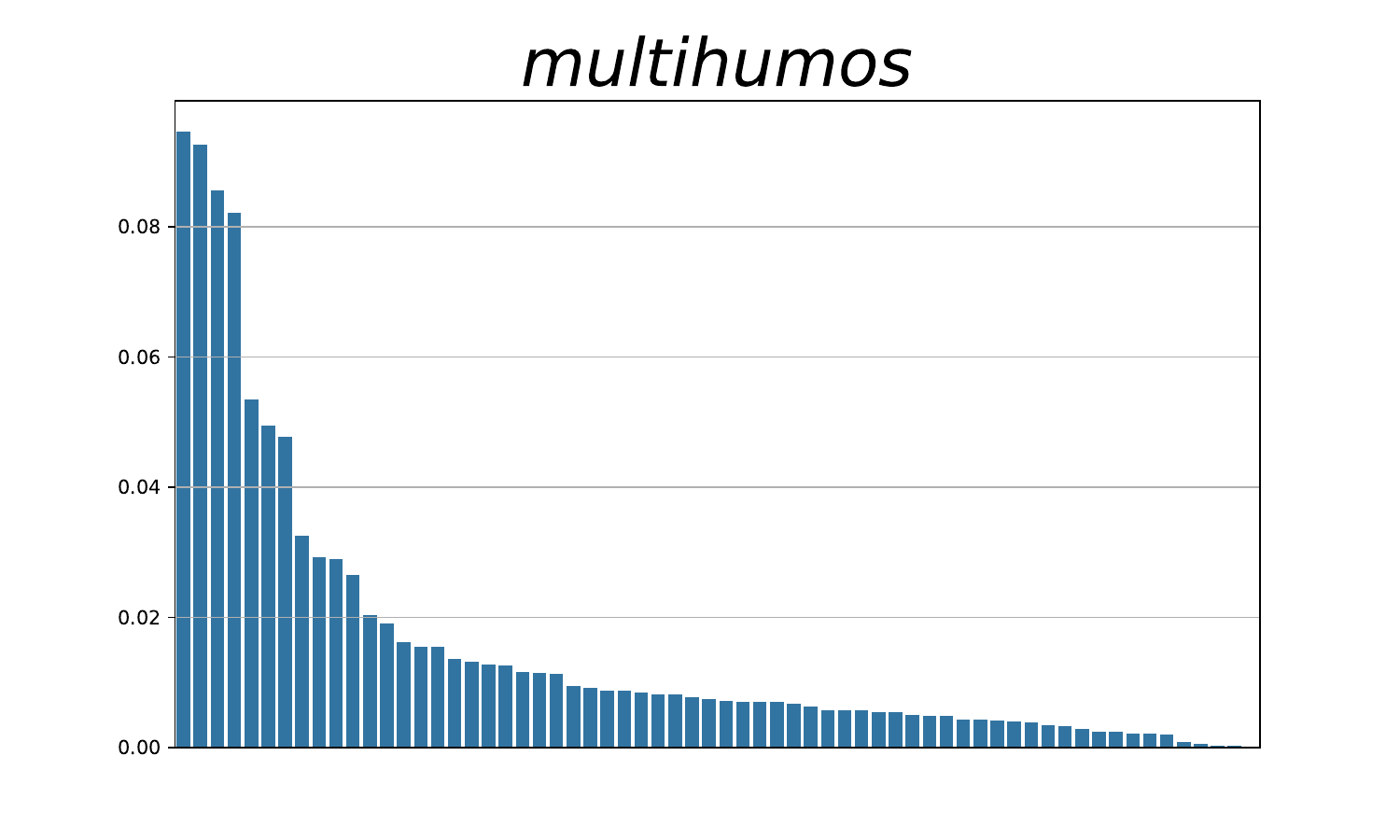}
    \end{minipage}
    
    % % if using 4, use this
    % \begin{minipage}[b]{0.45\textwidth}
    %     \includegraphics[width=\textwidth]{figures/dataset_skew_plots/epic_kitchen.pdf}
    % \end{minipge}
    % \begin{minipage}[b]{0.245\textwidth}
    %     \includegraphics[width=\textwidth]{figures/dataset_skew_plots/amazon.pdf}

    % \end{minipage}
    
    % \begin{minipage}[b]{0.245\textwidth}
    %     \includegraphics[width=\textwidth]{figures/dataset_skew_plots/retweet.pdf}

    % \end{minipage}
    % \begin{minipage}[b]{0.245\textwidth}
    %     \includegraphics[width=\textwidth]{figures/dataset_skew_plots/taxi.pdf}

    % \end{minipage}
    % \begin{minipage}[b]{0.245\textwidth}
    %     \includegraphics[width=\textwidth]{figures/dataset_skew_plots/taobao.pdf}

    % % if using 4, use this
    % \end{minipage}
    % \begin{minipage}[b]{0.45\textwidth}
    %     \includegraphics[width=\textwidth]{figures/dataset_skew_plots/stackoverflow.pdf}

    % \end{minipage}
    
\caption{Normalized event counts (y-axis) vs. event types sorted by count (x-axis) for two datasets - Breakfast and MultiTHUMOS, showing significant class imbalance.}

\label{fig:class_imbalance}
\end{figure}

We observe significant class imbalance in our datasets, as shown in \autoref{fig:class_imbalance} for the Breakfast and MultiTHUMOS datasets. This imbalance motivates our choice of Macro-F1 as the primary metric, as it treats all classes equally, unlike Accuracy, which is heavily influenced by the dominant class.

\subsection{LASTS representation of Asynchronous time series for Zero Shot}
\label{sec:zero_shot_prompts}
Here we present the LASTS prompt structure for use with LLMs for various tasks. The structure of the LASTS prompts is shown in \autoref{fig:last_prompt}.

\paragraph{System Prompt} The system prompt is very similar across tasks, except for the task specific portions of the prompt.  The system prompt used for Forecasting is: 
\begin{tcolorbox}[colback=blue!10, colframe=black!80,]
You are a helpful assistant. Your task is to complete an asynchronous time series. \textit{\textcolor{red}{dataset\_description}}. Each series is given in the format (inter\_arrival\_time, action\_name). This indicates that the action\_name started inter\_arrival\_times milliseconds after the start of the previous action or the beginning of time if it's the first action. The allowable actions are: \textit{\textcolor{red}{valid\_vocab}}. Given the first few elements of an asynchronous time series, your task is to provide  the next action with its inter arrival time as (inter\_arrival\_time, action\_name). You generate all your response as a single python tuple. Be sure to provide only that one python tuple and nothing else.
\end{tcolorbox}

The system prompt used for Imputation is:
\begin{tcolorbox}[colback=blue!10, colframe=black!80,]
You are a helpful assistant. Your task is to find a missing value in an asynchronous time series. \textit{\textcolor{red}{dataset\_description}}. Each series is given in the format (inter\_arrival\_time, action\_name). This indicates that the action\_name started inter\_arrival\_times milliseconds after the start of the previous action or the beginning of time if it's the first action. The allowable actions are: \textit{\textcolor{red}{valid\_vocab}}. One of the elements in the series would be missing, marked by the word 'MISSING'. Provide your answer as a single python tuple (inter\_arrival\_time, action\_name) which is your estimate of the missing element of the series. Be sure to  give me that one missing  python tuple as your response and nothing else.
\end{tcolorbox}
The system prompt for Anomaly Detection is:
\begin{tcolorbox}[colback=blue!10, colframe=black!80,]
You are a helpful assistant. Your task is to find an anomolous value in an asynchronous time series. \textit{\textcolor{red}{dataset\_description}}. Each series is given in the format (inter\_arrival\_time, action\_name). This indicates that the action\_name started inter\_arrival\_times milliseconds after the start of the previous action or the beginning of time if it's the first action. The allowable actions are: \textit{\textcolor{red}{valid\_vocab}}. One of the elements in the series is an anomaly, and your task is to identify this element which doesn't belong in the series. Provide your answer as a single python tuple (inter\_arrival\_time, action\_name) which is an element from the series you think is an anomaly. Just give me that one anomolous python tuple from the series as your answer and nothing else. 
\end{tcolorbox}

Here, \textbf{dataset\_description} is a short one line description of the underlying dataset, for example:  \textit{"The underlying dataset is derived from tagged human actions while cooking/preparing meals"}.

Also, \textbf{valid\_vocab} is a comma separated list of allowable action descriptions, if we choose to provide this list and if this list is small.

\paragraph{User Prompt} The user prompt in all three tasks is a comma separated string of sequence events, for example 
\begin{equation*}
    \textit{(0,wait),(139000,carry\_bowl),(26000,hold\_bowl),}
\end{equation*}
In case of imputation, there would be a missing element marked by the word MISSING, like so:
\begin{equation*}
    \textit{(0,wait),(139000,carry\_bowl),\textcolor{red}{MISSING},(41000,reach\_eggcarton),}
\end{equation*}

\paragraph{Assistant Prompt} This is empty for zero-shot, as it is filled by the LLM as its prediction for the task on the given sequence.

\subsection{Evaluating LLM Interaction with LASTS Components}
\label{ast_rep_zero_shot}

We considered various variants of framing the LASTS prompt and present a few interesting ones here, evaluated on Breakfast dataset. 

\paragraph{Testing LLMs use of world knowledge} We want to test whether LLMs can understand a prompt like LASTS and provide a meaningful response to the task on the sequence using their world knowledge. To this end, we study a variant where each event description is replaced by a uniquely mapped gibberish 4-letter string. This unique mapping ensures that while any semantic meaning in the descriptions is removed, the structure of the time series remains intact.\autoref{tab:scrambled_names} shows that all tracked metrics degrade considerably in the scrambled names variant. This confirms that LLMs  not only understand LASTS properly but also leverage their world knowledge to perform the specified tasks.

\begin{table}[ht]
\centering
\begin{tabular}{l r r r r r r}
\hline
                & \multicolumn{6}{c}{\textbf{Forecast}} \\ \hline
                & \textbf{M-F1} $\uparrow$ & \multicolumn{1}{c}{\% $\Delta$} & \textbf{Acc} $\uparrow$& \multicolumn{1}{c}{\% $\Delta$} & \textbf{MAE $\downarrow$} & \multicolumn{1}{c}{\% $\Delta$} \\ \hline
Zero Shot       & 0.0432 &       & 0.0866 &       & 37.8030 &       \\
Scrambled Names & 0.0140 & \textcolor{red}{$\downarrow$ -67.63\%} & 0.0397 & \textcolor{red}{$\downarrow$ -54.13\%} & 38.0742 & \textcolor{red}{$\uparrow$ 0.72\%} \\ \hline
                & \multicolumn{6}{c}{\textbf{Imputation}} \\ \hline
Zero Shot       & 0.0248 &       & 0.0338 &       & 33.7669 &       \\
Scrambled Names & 0.0100 & \textcolor{red}{$\downarrow$ -59.73\%} & 0.0224 & \textcolor{red}{$\downarrow$ -33.73\%} & 40.4918 & \textcolor{red}{$\uparrow$ 19.92\%} \\ \hline
                & \multicolumn{6}{c}{\textbf{Anomaly Detection}} \\ \hline
Zero Shot       & 0.0760 &       & 0.0650 &       & NA      &       \\
Scrambled Names & 0.0619 & \textcolor{red}{$\downarrow$ -18.55\%} & 0.0469 & \textcolor{red}{$\downarrow$ -27.88\%} & NA      &       \\ \hline
\end{tabular}
\caption{Comparing LASTS Zero Shot with the Scrambled Names variant across Forecast, Imputation, and Anomaly Detection tasks. Higher values are better for M-F1 and Acc, while lower values are better for MAE. Red indicates negative impact, while green indicates favorable impact.}
\label{tab:scrambled_names}
\end{table}
\definecolor{darkgreen}{rgb}{0.0, 0.5, 0.0}

\paragraph{Sequence Representation} We probe about the right representation for the time series events - should they be represented as $(e_i, t_i)$ or $(t_i, e_i)$. Our results in \autoref{tab:seq_rep} show that its better to have time first, followed by the event description. This is what we adopt in LASTS.
\begin{table}[ht]
\centering
\begin{tabular}{l r r r r r r}
\hline
                & \multicolumn{6}{c}{\textbf{Forecast}} \\ \hline
                & \textbf{M-F1} $\uparrow$ & \multicolumn{1}{c}{\% $\Delta$} & \textbf{Acc} $\uparrow$& \multicolumn{1}{c}{\% $\Delta$} & \textbf{MAE $\downarrow$} & \multicolumn{1}{c}{\% $\Delta$} \\ \hline
Time First  $(t_i, e_i)$     & 0.0432 &       & 0.0866 &       & 37.8030 &       \\
Event First $(e_i, t_i)$ & 0.0409 & \textcolor{red}{$\downarrow$ 5.38\%} & 0.0726 & \textcolor{red}{$\downarrow$ 16.07\%} & 37.5344 & \textcolor{darkgreen}{$\downarrow$ 0.71\%} \\ \hline
                & \multicolumn{6}{c}{\textbf{Imputation}} \\ \hline
Time First  $(t_i, e_i)$       & 0.0248 &       & 0.0338 &       & 33.7669 &       \\
Event First $(e_i, t_i)$ & 0.0071 & \textcolor{red}{$\downarrow$ -71.30\%} & 0.0150 & \textcolor{red}{$\downarrow$ -55.56\%} & 31.8194 & \textcolor{darkgreen}{$\downarrow$ -5.77\%} \\ \hline
                & \multicolumn{6}{c}{\textbf{Anomaly Detection}} \\ \hline
Time First  $(t_i, e_i)$       & 0.0760 &       & 0.0650 &       & NA      &       \\
Event First $(e_i, t_i)$ & 0.0858 & \textcolor{darkgreen}{$\uparrow$ 12.94\%} & 0.0619 & \textcolor{red}{$\downarrow$ -4.81\%} & NA      &       \\ \hline
\end{tabular}
\caption{Comparison of two ways to express events in an asynchronous time series - event first or time first across Forecast, Imputation, and Anomaly Detection tasks. Higher values are better for M-F1 and Acc, while lower values are better for MAE. Red indicates negative impact, while green indicates favorable impact.}
\label{tab:seq_rep}
\end{table}

\paragraph{Time Representation} We investigate if simplifying the series representation would improve LLM performance. For the Breakfast dataset, we replace inter-arrival times with durations, since we hypothesize that most actions occur contiguously for this dataset. We hypothesize that durations may be easier for the LLM to model rather than inter arrival. From the results in \autoref{tab:durations}, we observe that while we have a favourable impact on forecast, both imputation and anomaly detection suffer from this change. This suggests that while durations help with forecasting, more precise inter-arrival times are crucial for more involved tasks like imputation and anomaly detection. 
\begin{table}[ht]
\centering
\begin{tabular}{l r r r r r r}
\hline
                & \multicolumn{6}{c}{\textbf{Forecast}} \\ \hline
                & \textbf{M-F1} $\uparrow$ & \multicolumn{1}{c}{\% $\Delta$} & \textbf{Acc} $\uparrow$& \multicolumn{1}{c}{\% $\Delta$} & \textbf{MAE $\downarrow$} & \multicolumn{1}{c}{\% $\Delta$} \\ \hline
Zero Shot       & 0.0432 &       & 0.0866 &       & 37.8030 &       \\
Durations & 0.0600 & \textcolor{darkgreen}{$\uparrow$ 38.84\%} & 0.0953 & \textcolor{darkgreen}{$\uparrow$ 10.12\%} & 33.781 & \textcolor{darkgreen}{$\downarrow$ 10.62\%} \\ \hline
                & \multicolumn{6}{c}{\textbf{Imputation}} \\ \hline
Zero Shot       & 0.0248 &       & 0.0338 &       & 33.7669 &       \\
Durations & 0.0140 & \textcolor{red}{$\downarrow$ -43.56\%} & 0.0288 & \textcolor{red}{$\downarrow$ -14.81\%} & 29.6881 & \textcolor{darkgreen}{$\downarrow$ -12.09\%} \\ \hline
                & \multicolumn{6}{c}{\textbf{Anomaly Detection}} \\ \hline
Zero Shot       & 0.0760 &       & 0.0650 &       & NA      &       \\
Durations & 0.0767 & \textcolor{darkgreen}{$\uparrow$ 0.96\%} & 0.0532 & \textcolor{red}{$\downarrow$ -18.20\%} & NA      &       \\ \hline
\end{tabular}
\caption{Comparison of LASTS Zero Shot with the variant using durations instead of inter-arrival times across Forecast, Imputation, and Anomaly Detection tasks. Higher values are better for M-F1 and Acc, while lower values are better for MAE. Red indicates negative impact, while green indicates favorable impact.}
\label{tab:durations}
\end{table}

\subsection{LASTS representation used for LLM Adaptation}
\label{sec:lasts_for_peft}
For our experients on LLM adaptation, we keep the LASTS representation very similar to our zero shot experiments: 
\begin{itemize}
    \item \textbf{System prompt} in this case is a very concise description of just the task. We skip any dataset description as we expect the model to learn that during the fine tuning process.
    \item \textbf{User prompt} is represented as a comma separated sequence of tuples of event description and inter arrival times.
    \item \textbf{Assistant prompt}  contains the expected prediction.
\end{itemize}

The exact system prompt used for each of the tasks are as follows:
\begin{itemize}
    \item \textbf{Forecasting}: \textit{"Predict the next element of this asynchronous time series where each element is of the form (inter\_arrival\_time,\ action\_name)."}
    \item \textbf{Imputation}: \textit{"Predict the  element marked 'MISSING' in this asynchronous time series where each element is of the form (inter\_arrival\_time, action\_name)."}
    \item \textbf{Anomaly Detection}: \textit{"One of the element in this asynchronous time series is anomalous, find this element. Each element of the series is of the form (inter\_arrival\_time, action\_name)."}
\end{itemize}

\subsection{Baselines}
\label{sec:llms_for_ts_baselines}

\paragraph{Random Baseline}  To evaluate our methods on the three text-based datasets and the three tasks, we establish a random baseline simulating random guesses. For forecasting and imputation, given an input asynchronous time series, the baseline predicts the inter-arrival time as the average of all inter-arrival times in the sequence and selects a random event type from the valid event descriptions. For anomaly detection, it randomly labels an event from the series as anomalous (see \autoref{tab:performance_comparison}).

\paragraph{Foundation Models for Time Series Baseline}
We adapted Chronos \citep{ansari2024chronos}, a state-of-the-art foundation model designed for zero-shot forecasting on time series data, as a baseline for forecasting and imputation tasks on asynchronous time series datasets. We use the largest model version (\textit{amazon/chronos-t5-large}) available which contains $710M$ model parameters. Since Chronos exclusively handles numerical data, we converted our event descriptions into categorical representations. Each asynchronous time series of length $n$ was transformed into a sequence of $2n$ integers, alternating between inter-arrival times and event categories.

For forecasting, the task was framed as predicting the next two elements in this sequence given the historical context. Adapting Chronos for imputation, however, required additional considerations since it is inherently designed for forecasting. We reformulated the imputation task as a forecasting problem: if the prefix leading up to the missing element is longer than the suffix following it, we treated imputation as forecasting the missing element using the prefix as context. Conversely, if the suffix is longer, we reversed the suffix and used it as context to forecast the missing element. This approach ensures the longest possible context is utilized for predicting the missing value.

It is worth noting that adapting Chronos for anomaly detection is not straightforward, as anomaly detection involves identifying a single anomalous event within the series, which does not align with Chronos' forecasting capabilities. Consequently, Chronos is provided as a baseline exclusively for forecasting and imputation tasks. 

\paragraph{LLMs for Time Series Baselines}
We adapted two LLM-based methods for time series: \textbf{LLMTime} \citep{gruver2024large} and \textbf{LLMProcesses} \citep{requeima2024llm}, as baselines. Since both methods are designed for numerical time series, we converted textual event descriptions into categorical representations.

\subparagraph{LLMTime}
In this method, each data point is represented as a pair: (inter-arrival-time, event-categorical). We modified the default next-token prediction behavior of the model using simple task-specific prompts:
\begin{itemize} \item \textbf{Forecasting}: \textit{Predict the next time and event.} \item \textbf{Imputation}: \textit{Find the element marked as 'MISSING.'} \item \textbf{Anomaly Detection}: \textit{Find the anomalous time and event.} \end{itemize}

\subparagraph{LLMProcesses}
This method uses in-context learning with $(x, y)$ examples derived from a sequence, treating the sequence as a real-valued function on a 2D space as domain. In this setup, $x$ represents a point in 2D space $(x_1, x_2)$, where $x_1$ denotes the sequence position, and $x_2$ indicates the output type: $0$ for inter-arrival time and $1$ for event categorical. For a given sequence, we crafted two distinct prompts: one for predicting the event categorical and another for predicting the inter-arrival time, based on the corresponding value of $x$. We followed the recommended settings from the original paper for prompt construction.

 However, anomaly detection does not align with this framework, as it involves identifying a single anomalous time point where the function output is $0$ everywhere except at the anomaly. This makes it unsuitable for predicting function values at unseen points based on prior observations. Consequently, we adapted this approach exclusively for forecasting and imputation tasks.

% We use two LLM-based time series forecasting methods, LLMTime \citep{Gruver2023Large} and LLMProcesses \citep{requeima2024llm}, as baselines for asynchronous time series. Since LLMProcesses cannot be easily adapted for anomaly detection, it is used only for forecasting and imputation. Our method performs comparably or better than these baselines. Details on baseline implementation are in Appendix 
\paragraph{TPP Models as Baselines}
We compare our best fine-tuned model configuration, $LASTS+StoP$, against current state-of-the-art methods for forecasting on asynchronous time series. These methods are adapted from the benchmark study in \citep{Xue2023EasyTPP}. The evaluation spans eight datasets, five of which—Amazon, Retweet, Taxi, Taobao, and StackOverflow  contain event categoricals without textual descriptions and are regarded as standard benchmarks for asynchronous time series analysis.

We benchmark the TPP models covered in the EasyTPP benchmark \citep{Xue2023EasyTPP} on the three textual datasets considered in our work: Breakfast, MultiTHUMOS, and EPIC KITCHEN. Since these datasets represent events as text and TPP models are not equipped to handle text directly, we converted the event names into event categoricals to make them compatible with these models.

\begin{figure}[h!]
    \centering
    \includegraphics[width=\textwidth]{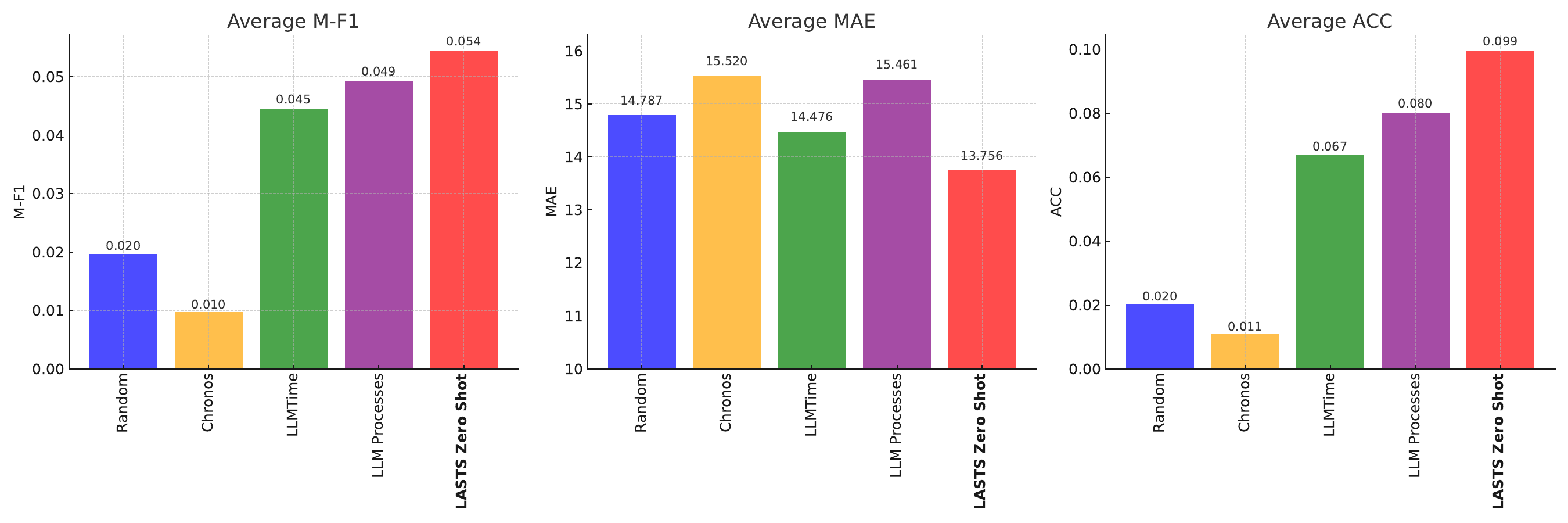}
    \caption{
        Comparison of performance metrics: Macro-F1 (M-F1), Mean Absolute Error (MAE), and Accuracy (ACC), averaged across all datasets for Forecast and Imputation tasks. Higher values for M-F1 and ACC indicate better performance, while a lower value of MAE is preferred. It is evident that LASTS Zero Shot (our method) achieves the highest average M-F1 and average ACC among all the baselines and also produces the lowest MAE.
    }
    \label{fig:baseline_performance_plot}
\end{figure}

\paragraph{Observations} We summarize our comparison of various baselines with LASTS Zero Shot in \autoref{fig:baseline_performance_plot}. We observe that Chronos performs the weakest among the baselines, yet it remains competitive. This is expected as Chronos, while being a much smaller model compared to LLMs, is highly specialized for time series forecasting, which enables it to achieve decent performance. LLMTime and LLMProcesses also perform competitively, especially on the MultiTHUMOS dataset. We attribute this to the noisy nature of the MultiTHUMOS dataset, which includes non-standard event names (e.g., "OneHandedCatch," "TalkToCamera", etc) and repetitive, less meaningful patterns (e.g., "GolfSwing, Wait, GolfSwing, Wait..."). These characteristics may help event-categorical-based models like LLMTime and LLMProcesses. However, on the other two datasets—Breakfast and EPIC\_KITCHEN—the textual descriptions of events provide a significant advantage, as evident from the comfortable margin by which LASTS Zero Shot outperforms LLMTime and LLMProcesses across all tasks.

Furthermore, we observed that existing TPP-based models struggled with datasets containing a large number of unique event types, often performing poorly, failing to converge, or encountering out-of-memory errors. This highlights the challenges these models face in handling the diversity and complexity of such datasets.

\subsection{Disentangling Stochasticity and Prefix Picking in StoP}
\label{app:stop_disentangling}

\begin{figure}[]
    \centering
    \begin{minipage}[t]{0.49\columnwidth}
        \centering
        \includegraphics[width=\linewidth]{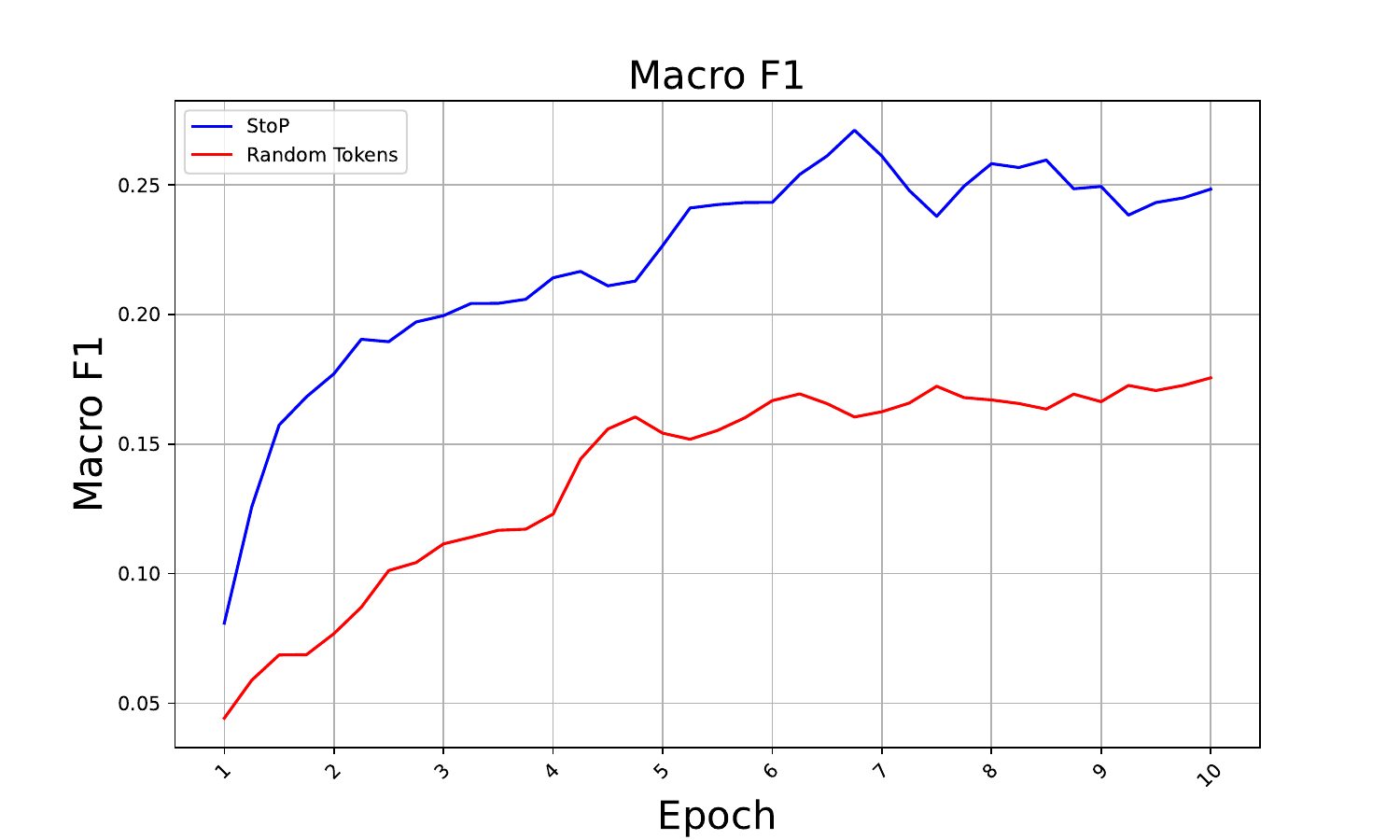}
    \end{minipage}
    \hfill
    \begin{minipage}[t]{0.49\columnwidth}
        \centering
        \includegraphics[width=\linewidth]{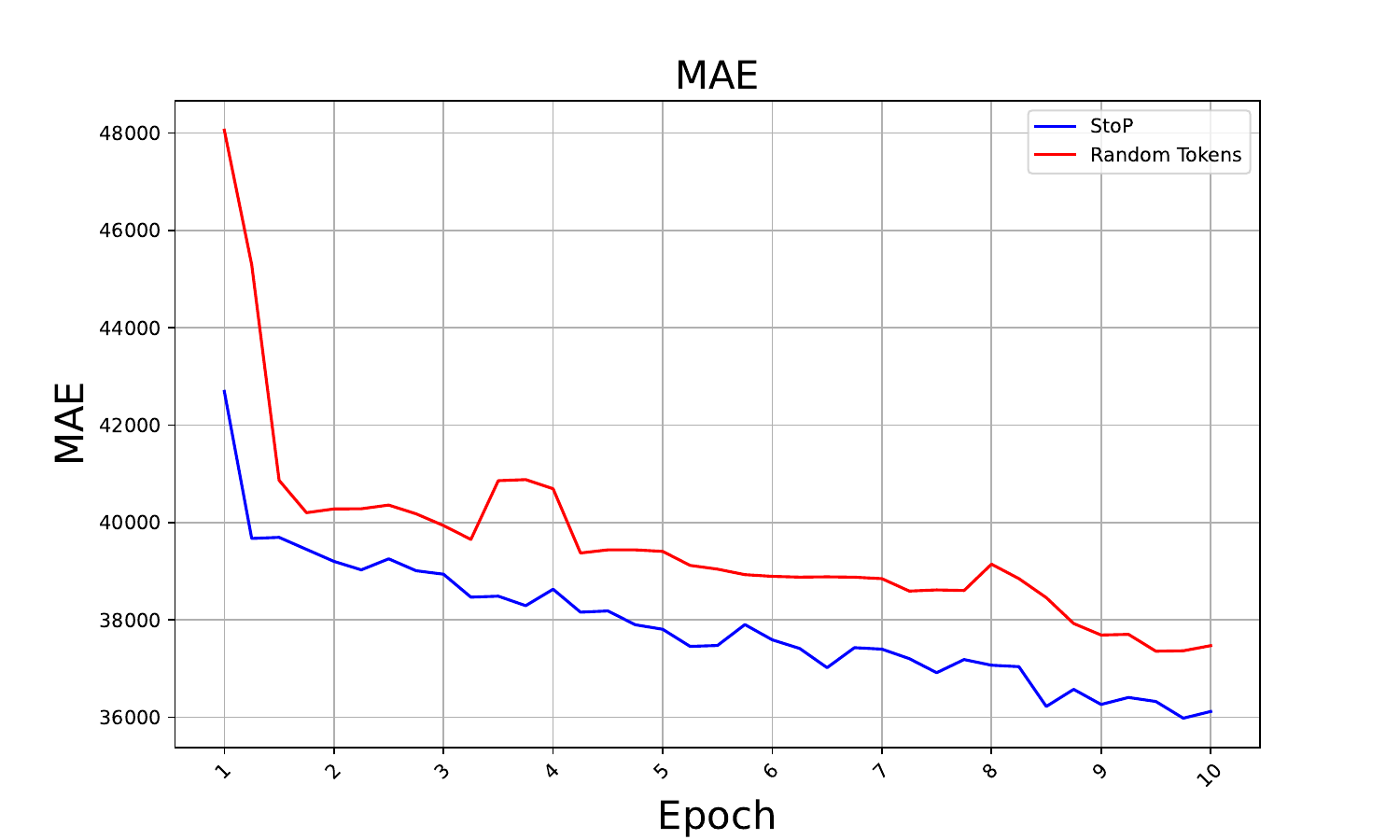}
    \end{minipage}
    
    \caption{Comparison of Macro-F1 and MAE for StoP vs. random token selection during training, evaluated on validation data after 10 epochs. Results show that random token selection (shown in red) fails to learn effective prompts, while StoP's structured prefix selection (shown in blue) achieves significantly better performance.}
    \label{fig:random_tokens}
\end{figure}

To further analyze the impact of prefix picking in StoP, we compare it with an alternative training paradigm where, instead of selecting a structured prefix, we randomly select $l$ tokens from the prompt during each batch, with $l$ drawn from a uniform distribution. This comparison isolates the effects of introducing stochasticity alone versus the structured prefix picking employed by StoP. \autoref{fig:random_tokens} presents a comparison of Macro-F1 and MAE metrics on the validation set as both prompts are trained for 10 epochs.  These plots show that stochasticity alone is not sufficient for learning good soft prompts, and structured prefix picking is a key component of the StoP training. 

\subsection{StoP Prompt Interpretations Through Model Probing} 
\label{sec:prompt_interpretations_word_embeddings}
Prior work such as \cite{Lester2021Power} attempts to interpret learned prompts by mapping them to the closest input embeddings. However, this approach often produces sequences that are difficult to understand or lack meaningful content. For example, \cite{xusoft} observed that the closest words to the learned embeddings are mostly meaningless, several tokens are mapped to the same word, and the cosine similarity between the tokens and their closest word embeddings almost always falls below $0.16$. Instead, we explore probing the LLM using the learned prompt. By appending the learned prompt with a simple instruction, such as \textit{"Tell me in as much detail as possible what task you are supposed to do,"} we encourage the LLM to generate an output that reflects its understanding of the task. This approach allows us to gain meaningful insights into what the model has summarized from the tasks and datasets it has been trained on. Here we present a few interesting interpretations of StoP prompts obtained by model probing.

\paragraph{StoP of Length $400$, Breakfast Dataset, Forecast } We obtain a very nice summarization of the breakfast dataset: 

\begin{tcolorbox}[colback=blue!10, colframe=black!80]
The task appears to be related to preparing a snack or meal by using the facilities provided in
the kitchen. Specifically, I was asked to perform a series of actions to accomplish this goal.
Here’s what I’ve observed:1. **Opening the cabinet**: I need to open the cabinet to access
the necessary utensils, cookware, or ingredients.2. **Grabbing the ingredients**: I will
need to pick up the specified ingredients, such as a bowl, a cutting board, or a can opener, to
prepare the desired dish.3. **Slicing the ingredients**: I will cut the ingredients into the re-
quired proportions, shapes, or sizes depending on the recipe being followed.4. **Preheating
the cooking surface**: If needed, I will turn on the stove, oven, or other cooking device to
the desired temperature to start cooking the ingredients.5. **Cooking the ingredients**: I
will proceed to cook the ingredients according to the required cooking method, such as boil-
ing, saut´eing, frying, or baking.6. **Adding condiments or seasonings**: If necessary, I
will add the required condiments, spices, or seasonings to the cooked ingredients to enhance
their flavor.7. **Combining the ingredients**: Finally, I will combine the cooked ingredi-
ents in the desired manner, whether it’s mixing, blending, or assembling the final dish.Please
let me know if I’m correct in my understanding or if there’s anything else I can do to help!
\end{tcolorbox}

\paragraph{StoP of Length $5$, Breakfast Dataset, Forecast } We obtain very cooking adjacent prompt interpretations from even a small prompt of length $5$ trained for forecasting on Breakfast. For example:

\begin{tcolorbox}[colback=blue!10, colframe=black!80]
Here is my main task: "Reach for a tablespoon and move in the direction of a plate where I already placed some eggs and a knife. Slowly pour one egg into the plate, take the knife, and carefully cut the egg in halves. Transfer both halves away from the plate, and then carry the remaining plate towards a sink. Turn the tap on, wash the plate, and then move away from the sink, holding the plate still under running water. Remove the plate from the sink, carry it to a countertop, and put it down. Finally, reach for a knife and move towards a bowl containing a knife, carry the knife, and move towards a counter where I already placed a cutting board, where there is a set of utensils lying on it. Take the knife, push towards the board, and reach for a knife. Move away from the cutting board.
\end{tcolorbox}

\paragraph{StoP of length $400$, MultiTHUMOS Dataset, Imputation} The interpretation seems to have mapped the idea of an asynchronous time series to another domain all together, while still keeping elements like walk, run, stand, etc from the original dataset:
\begin{tcolorbox}[colback=blue!10, colframe=black!80]
I'm an AI designed to generate text based on the sequence of events occurring in the events diagram. Specifically, I\'m trained to analyze the sequence of events and identify the corresponding actions or actions sequences in the diagram, and then use that information to generate a textual description of what is happening in the diagram.The main tasks I was designed to perform are:1. **Identify sequence of events**: I analyze the sequence of events in the diagram, identifying the time-stamped data points and the transitions between them.2. **Match events with actions**: I match the identified sequence of events with the corresponding actions or actions sequences. For example, if I see a sequence of time-stamped data points indicating a person sitting, standing, and then walking, I would match those events to the action of "walking".3. **Generate textual description**: Using the matched events and actions, I then generate a textual description of what is happening in the diagram. For example, if I match the sequence "sitting, standing, walking" to the action "walking", I would generate a text describing that sequence as "walking". Additionally, I also make use of some contextual information, such as the diagram\'s caption, to disambiguate and refine my generated textual descriptions. The final textual output is a natural language description of the content of the diagram.
\end{tcolorbox}

\paragraph{StoP of length $400$, MultiTHUMOS Dataset, Anomaly Detection} The interpretations center around games, which is expected as MultiTHUMOS is a gaming dataset. Here are two sampled interpretations:
\begin{tcolorbox}[colback=blue!10, colframe=black!80]
The task is to simulate a volleyball game. The events include:1. Slices or spikes the ball with the palm of the hand (VolleyballSpiking). 2. Blocks the ball with the palm of the hand (VolleyballBlock).3. Sets the ball with the palm of the hand (VolleyballSet). 4. Pounds or slams the ball with the fist (VolleyballSpiking). 5. Jumps or reaches for the ball (VolleyballJumping). 6. Moves sideways or slides on the ground (VolleyballSpikeRun).It's important to remember that volleyball spiking, blocking, setting, and jumping involves movement and body language, which can't be reflected directly in a sequential diagram
\end{tcolorbox}

\begin{tcolorbox}[colback=blue!10, colframe=black!80]
The task you are supposed to do is sports broadcasting, and specifically, you are a golf broadcaster. As a golfer, you are supposed to perform a live broadcast of the golf match, providing commentary and analysis as the golfers participate in the match. The situation is simulated, as there are no external stimuli or real golfers, you have to act as both commentator and golfer, making mock commentary and mimicking golf swings.The task requires you to stay in a fixed position, usually behind a microphone, and maintain a semi-serious, professional tone. Your task is to articulate the events on the course, such as drives, putts, and other golf shots, as if you are actually seeing them happen. You should also describe the terrain, distance, and obstacles faced by the golfer, and offer your expert analysis and opinions on their technique, strategy, and chances of scoring.Your commentary should be smooth, clear, and engaging, drawing the audience into the game and painting a vivid picture of the action on the course. Your performance should exude confidence and authority, as if you are an experienced commentator who has watched many matches before. The audience should feel like they are right there on the course with you, experiencing the excitement and suspense of the tournament.
\end{tcolorbox}

% \textbf{StoP of Length 400, Breakfast Dataset, Forecast} 

% \textit{Prompt:} Write a 5 line poem about your task. \\
% \textit{Model Response}: \\
% \begin{tcolorbox}[colback=blue!10, colframe=black!80, verbatim]
% Whisk, bowl, and cake, a trio so fine \\
% Cream and sugar, a dancer's rhyme \\ 
% Lather and smooth, a frosting divine \\ 
% Sweet delight, a treat so sublime \\ 
% Tartan, the crown, a masterpiece of mine. 
% \end{tcolorbox}

% \begin{tcolorbox}[colback=blue!10, colframe=black!80, verbatim]
% Topping pizza, a careful art \\
% Fresh herbs sprinkle, a savory heart \\
% Fragrance fills, the evening air \\
% Tempting taste buds, beyond compare \\ 
% A culinary delight, beyond despair 
% \end{tcolorbox}

% \begin{tcolorbox}[colback=blue!10, colframe=black!80, verbatim]
% Within the realm of fruitful zest \\
% Wielding knife and gentle quest \\ 
% With each delicate cut, I unfold \\ 
% A tale of taste and weetness to hold \\ 
% And savor the beauty within 
% \end{tcolorbox}

\textbf{StoP of Length 400, Taobao Dataset, Forecast} Note that this dataset is not textual, so all the soft prompt has seen is the task description and a sequence of pairs of numbers.

\textit{Prompt:} Describe your task. \\
\textit{Model Response}: \\

\begin{tcolorbox}[colback=blue!10, colframe=black!80]
I\'m an AI assistant, my task is to generate a sequence based on a rule that I\'ve learned 
from given inputs (or "train" sequences). The rule is inferred by analyzing the pattern of increasing or decreasing values of the input sequence. Each sequence is either called ("silenced") or activated, and if there\'s no change in the sequence, the "silence" is used, unless the input sequence calls the "silence" by itself. The task involves predicting the next output in a sequence, given some input sequences that establish a pattern.
\end{tcolorbox}   

\subsection{Comparison of LASTS + StoP with other PEFT techniques}
\label{app:stop_comparison_with_peft}
In this section, we compare the performance of LASTS + StoP with other PEFT techniques listed in Table \ref{tab:performance_comparison}. Table \ref{tab:compare_stop_and_sp} highlights the percentage improvements observed in various metrics when using Stochastic Soft Prompting compared to standard Soft Prompting. We observe a significant advantage of Stochastic Soft Prompting across all datasets and tasks, with an overall average increase of $12.69\%$ in Macro-F1 across all tasks and datasets. Similarly, Table \ref{tab:compare_stop_and_qlora} demonstrates an average increase of $13.55\%$ in Macro-F1 when using Stochastic Soft Prompting instead of finetuning techniques like QLORA.

\begin{table}[ht]
\centering
\resizebox{\textwidth}{!}{%
\begin{tabular}{lccccccccc}
\toprule
\multirow{2}{*}{\textbf{Task}} & \multicolumn{3}{c}{\textbf{Breakfast}} & \multicolumn{3}{c}{\textbf{MultiTHUMOS}} & \multicolumn{3}{c}{\textbf{EPIC\_KITCHEN}} \\ 
\cmidrule(lr){2-4} \cmidrule(lr){5-7} \cmidrule(lr){8-10}
 & \textbf{M-F1} & \textbf{MAE} & \textbf{ACC} & \textbf{M-F1} & \textbf{MAE} & \textbf{ACC} & \textbf{M-F1} & \textbf{MAE} & \textbf{ACC} \\ 
\midrule
Forecast       & 11.09\% & 0.91\% & 4.87\% & 6.08\% & 0.77\% & 0.04\% & 2.13\% & -4.91\% & 3.52\% \\
Imputation     & 23.40\% & 2.22\% & 17.37\% & 7.64\% & 2.76\% & 10.95\% & 30.66\% & 1.09\% & 10.81\% \\
Anomaly Detection & 10.40\% & \textemdash     & 12.82\% & 15.56\% & \textemdash     & 10.97\% & 7.21\% & \textemdash     & 6.06\% \\
\midrule
\textbf{Avg Gain (Per Task)} & 14.96\% & 1.56\% & 11.69\% & 9.76\% & 1.76\% & 7.32\% & 13.33\% & -1.91\% & 6.80\% \\
\midrule
\textbf{Avg Gain (All Tasks, All Datasets)} & \multicolumn{3}{c}{\textbf{M-F1}: 12.69\%} & \multicolumn{3}{c}{ \textbf{MAE}: 0.47\%} & \multicolumn{3}{c}{ \textbf{ACC}: 8.60\%} \\
\bottomrule
\end{tabular}%
}
\caption{Comparison of LASTS+StoP with LASTS+SP. The table shows the percentage improvement in each metric achieved by using Stochastic Soft Prompting compared to standard Soft Prompting. Significant gains are observed across all datasets and tasks with Stochastic Soft Prompts. On average, across all datasets and tasks, Macro F1 increases by $12.69\%$.}  
\label{tab:compare_stop_and_sp}
\end{table}

\begin{table}[ht]
\centering
\resizebox{\textwidth}{!}{%
\begin{tabular}{lccccccccc}
\toprule
\multirow{2}{*}{\textbf{Task}} & \multicolumn{3}{c}{\textbf{Breakfast}} & \multicolumn{3}{c}{\textbf{MultiTHUMOS}} & \multicolumn{3}{c}{\textbf{EPIC\_KITCHEN}} \\ 
\cmidrule(lr){2-4} \cmidrule(lr){5-7} \cmidrule(lr){8-10}
 & \textbf{M-F1} & \textbf{MAE} & \textbf{ACC} & \textbf{M-F1} & \textbf{MAE} & \textbf{ACC} & \textbf{M-F1} & \textbf{MAE} & \textbf{ACC} \\ 
\midrule
Forecast       & 2.93\% & 4.39\% & 3.11\% & 22.65\% & 4.71\% & 10.31\% & 4.32\% & -4.47\% & 6.39\% \\
Imputation     & 22.27\% & 1.20\% & 9.60\% & 3.80\% & -5.40\% & 3.46\% & 61.38\% & 0.25\% & 25.24\% \\
Anomaly Detection & 2.67\% & \textemdash     & 3.40\% & 0.70\% & \textemdash     & 1.65\% & 1.27\% & \textemdash     & 0.70\% \\
\midrule
\textbf{Avg Gain (Per Task)} & 9.29\% & 2.79\% & 5.37\% & 9.05\% & -0.34\% & 5.14\% & 22.32\% & -2.11\% & 10.78\% \\
\midrule
\textbf{Avg Gain (All Tasks, All Datasets)} & \multicolumn{3}{c}{\textbf{M-F1}: 13.55\%} & \multicolumn{3}{c}{\textbf{MAE}: 0.11\%} & \multicolumn{3}{c}{\textbf{ACC}: 7.10\%} \\
\bottomrule
\end{tabular}%
}
\caption{Comparison of LASTS+StoP with LASTS+QLORA. The table shows the percentage improvement in each metric achieved by using Stochastic Soft Prompting compared to finetuning via QLORA. Significant gains are observed across all datasets and tasks with Stochastic Soft Prompts. On average, across all datasets and tasks, Macro-F1 increases by $13.55\%$.}
\label{tab:compare_stop_and_qlora}
\end{table}

% \usepackage{graphicx}
\begin{table}[H]
\centering
\resizebox{\textwidth}{!}{%
\begin{tabular}{|l|c|cc|cc|cc|}\toprule
 & 
  \multicolumn{1}{l|}{} &
  \multicolumn{2}{c|}{Breakfast} &
  \multicolumn{2}{c|}{MultiThumos} &
  \multicolumn{2}{c|}{EPIC KITCHEN} \\ \midrule
 &
  \# Params &
  \multicolumn{1}{c|}{Macro F1 $\uparrow$} &
  MAE $\downarrow$ &
  \multicolumn{1}{c|}{Macro F1 $\uparrow$} &
  MAE $\downarrow$ &
  \multicolumn{1}{c|}{Macro F1 $\uparrow$} &
  MAE $\downarrow$ \\ \midrule

\multicolumn{1}{|c|}{Forecast} &
  1B &
  \multicolumn{1}{c|}{0.2292} &
  33.9309 &
  \multicolumn{1}{c|}{0.3210} &
  1.8013 &
  \multicolumn{1}{c|}{0.0574} &
  3.0859 \\
 &
  3B &
  \multicolumn{1}{c|}{0.2526} &
  33.2541 &
  \multicolumn{1}{c|}{0.3694} &
  1.7259 &
  \multicolumn{1}{c|}{0.0708} &
  3.0169 \\ 
 &
  8B &
  \multicolumn{1}{c|}{0.2633} &
  32.5464 &
  \multicolumn{1}{c|}{0.3947} &
  1.6503 &
  \multicolumn{1}{c|}{0.0797} &
  3.0318 \\ \midrule

\multicolumn{1}{|c|}{Imputation} &
  1B &
  \multicolumn{1}{c|}{0.0256} &
  31.1075 &
  \multicolumn{1}{c|}{0.0907} &
  2.4256 &
  \multicolumn{1}{c|}{0.0102} &
  3.2571 \\
 &
  3B &
  \multicolumn{1}{c|}{0.0966} &
  31.1597 &
  \multicolumn{1}{c|}{0.1329} &
  2.3963 &
  \multicolumn{1}{c|}{0.0280} &
  3.1445 \\
 &
  8B &
  \multicolumn{1}{c|}{0.2064} &
  28.2251 &
  \multicolumn{1}{c|}{0.2213} &
  2.3445 &
  \multicolumn{1}{c|}{0.0610} &
  3.1116 \\ \midrule
 
\multicolumn{1}{|c|}{Anomaly Detection} &
  1B &
  \multicolumn{1}{c|}{0.0688} &
  \textemdash &
  \multicolumn{1}{c|}{0.0954} &
  \textemdash &
  \multicolumn{1}{c|}{0.0318} &
  \textemdash \\
 &
  3B &
  \multicolumn{1}{c|}{0.5726} &
  \textemdash &
  \multicolumn{1}{c|}{0.4777} &
  \textemdash &
  \multicolumn{1}{c|}{0.5793} &
  \textemdash \\
 &
  8B &
  \multicolumn{1}{c|}{0.7198} &
  \textemdash &
  \multicolumn{1}{c|}{0.6045} &
  \textemdash &
  \multicolumn{1}{c|}{0.6603} &
  \textemdash \\ \bottomrule
\end{tabular}%
}
\caption{Comparison of Macro-F1 and MAE across the Breakfast, MultiThumos, and EPIC\_KITCHENS datasets for forecasting, imputation, and anomaly detection as the number of model parameters varies. The results show that Macro-F1 consistently improves with increasing model size across all datasets and tasks. In most cases, MAE decreases as model size increases, confirming that larger models generally lead to better performance.}
\label{tab:scaling_laws}
\end{table}

\subsection{Scaling to different LLM backbone sizes}
\label{app:scaling_laws}
We trained Stochastic Soft Prompts (StoP) across different backbone sizes of large language models and observed consistent improvements in performance as the model size increased. Specifically, we conducted experiments using LLama3.2 models with 1B and 3B parameters, as well as the LLama3-8B Instruct model. These improvements were clear across the Breakfast, MultiThumos, and EPIC\_KITCHENS datasets and applied to all tasks - forecasting, imputation, and anomaly detection.

\begin{figure}[h!]
    \centering
    \includegraphics[width=0.8\textwidth]{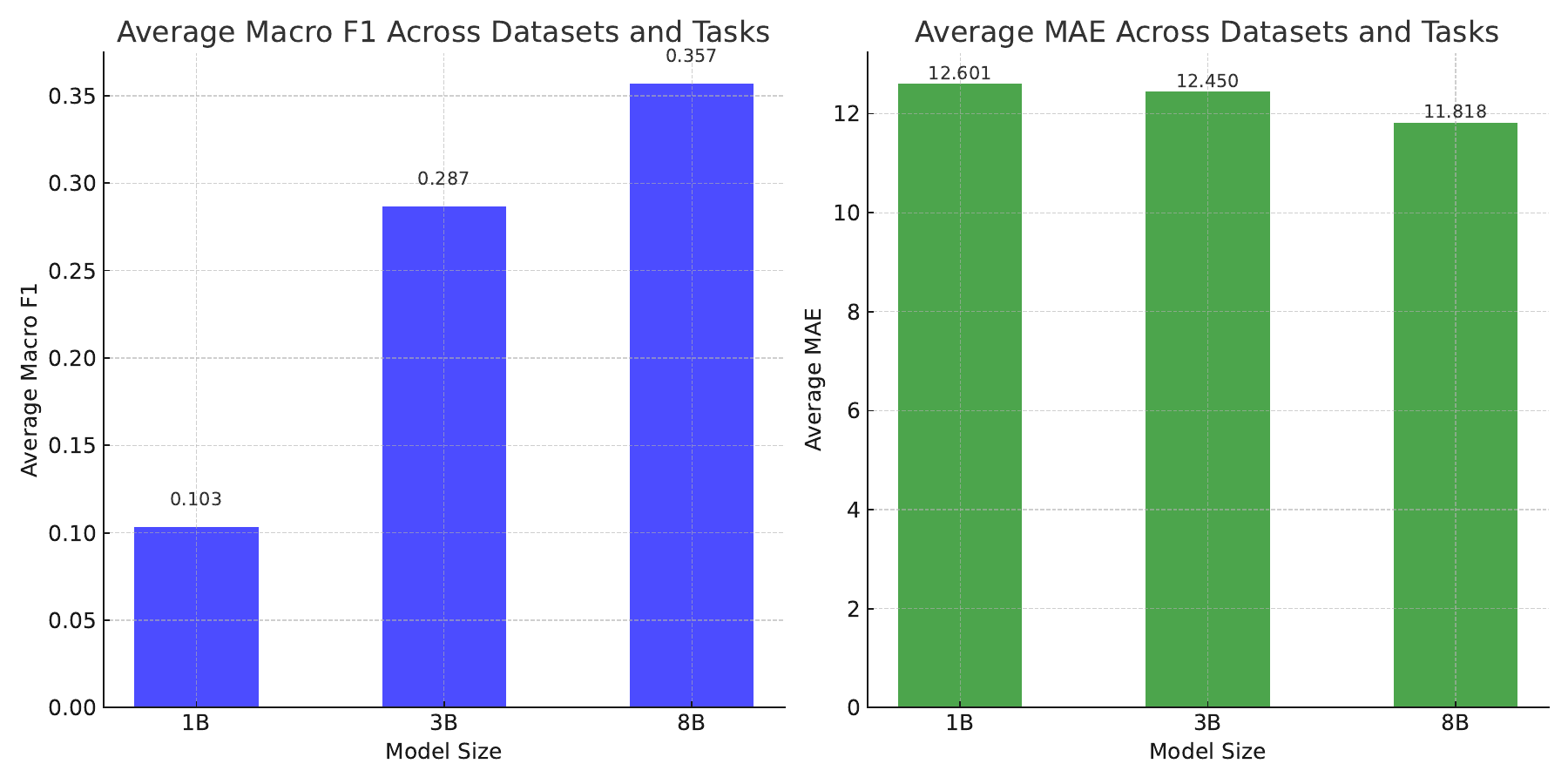}
    \caption{
        Comparison of average Macro F1 and MAE across all datasets and tasks for different model sizes. 
        The left histogram shows the average Macro F1 scores, while the right histogram depicts the average MAE values. 
        We see a clear trend of improvement in both metrics as model sizes increase.
    }
    \label{fig:avg_macro_f1_mae_by_model_size}
\end{figure}

Notably, \autoref{tab:scaling_laws} and \autoref{fig:avg_macro_f1_mae_by_model_size} show that macro-F1 scores consistently improve with larger model sizes across all datasets and tasks. Additionally, Mean Absolute Error (MAE) decreased in most cases as the model size increased, further confirming that larger models help Stochastic Soft Prompts perform better by utilizing their enhanced representational power. The performance difference between model sizes is smaller for forecasting tasks since these align with the next-token prediction that LLMs are trained on. However, for harder tasks like imputation and anomaly detection, the improvements are much larger as model size increases.

\begin{table}[H]
    \centering
    \resizebox{\textwidth}{!}{
    \begin{tabular}{l|ccc|ccc|ccc}
        \toprule
        \textbf{Few-Shot (k)} & \multicolumn{3}{c|}{\textbf{Breakfast}} & \multicolumn{3}{c|}{\textbf{MultiTHUMOS}} & \multicolumn{3}{c}{\textbf{EPIC-KITCHENS}} \\
        \midrule
        & M-F1 $\uparrow$ & MAE $\downarrow$  & ACC $\uparrow$  & M-F1 $\uparrow$  & MAE 
        $\downarrow$ & ACC $\uparrow$  & M-F1 $\uparrow$ & MAE $\downarrow$ & ACC $\uparrow$ \\
        \midrule
        & \multicolumn{9}{c}{Forecast}  \\
        \midrule
        \( k=0 \) & 0.0604 & 38.1630 & 0.0969 & 0.1361 & 1.8868 & 0.1826 & 0.0105 & 3.1566 & 0.0920 \\
        \( k=1 \) & 0.1312 & 37.6239 & 0.1808 & 0.1393 & 1.7913 & 0.2381 & 0.0144 & 3.2606 & 0.1123 \\
        \( k=2 \) & 0.1257 & 36.4688 & 0.1870 & 0.1622 & 1.7960 & 0.2505 &\textbf{ 0.0151} & 3.2266 & \textbf{0.1180 }\\
        \( k=5 \) & 0.1518 & \textbf{35.5605} & 0.2133 & 0.1676 & 1.8114 & 0.2581 & 0.0149 & 3.3092 & 0.1150 \\
        \( k=7 \) & 0.1491 & 35.6785 & 0.2107 & \textbf{0.1991} & \textbf{1.7810 }& \textbf{0.2828} & 0.0138 & 3.2177 & 0.1002 \\
        \( k=10 \) & \textbf{0.1667} & 37.6084 & \textbf{0.2442} & 0.1807 & 1.7820 & 0.2397 & 0.0124 & \textbf{3.0904} & 0.0901 \\
        \midrule
        & \multicolumn{9}{c}{Imputation}  \\
        \midrule
        \( k=0 \) & 0.0263 & 33.0097 & 0.0594 & 0.0915 & 2.6696 & 0.1210 & 0.0015 & 3.6527 & 0.0446 \\
        \( k=1 \) & 0.0419 & 33.1403 & 0.0738 & 0.1165 & 2.5106 & 0.1490 & 0.0018 & 3.6402 & \textbf{0.0569} \\
        \( k=2 \) & \textbf{0.0527} & \textbf{31.1138} & 0.0826 & 0.1102 & \textbf{2.3576} & 0.1486 & 0.0022 & 3.5375 & 0.0527 \\
        \( k=5 \) & 0.0520 & 33.3440 & 0.1001 & 0.1013 & 2.3982 & \textbf{0.1569} & \textbf{0.0023} & \textbf{3.2528} & 0.0547 \\
        \( k=7 \) & 0.0509 & 34.0198 & 0.0994 & 0.1001 & 2.4228 & 0.1462 & 0.0019 & 3.3447 & 0.0475 \\
        \( k=10 \) & 0.0474 & 31.2001 & \textbf{0.1069} & \textbf{0.1219} & 2.3771 & 0.1546 & 0.0015 & 3.2552 & 0.0406 \\
        \midrule
        & \multicolumn{9}{c}{Anomaly Detection}  \\
        \midrule
        \( k=0 \) & 0.0923 & \textemdash & \textbf{0.0763} & 0.2755 & \textemdash & 0.1949 & 0.0159 & \textemdash & 0.0777 \\
        \( k=1 \) & 0.1002 & \textemdash & 0.0681 & 0.2809 & \textemdash & 0.1961 & 0.0172 & \textemdash & 0.0854 \\
        \( k=2 \) & 0.0739 & \textemdash & 0.0569 & 0.3361 & \textemdash & \textbf{0.2891} & 0.0213 & \textemdash & 0.1062 \\
        \( k=5 \) & 0.0837 & \textemdash & 0.0563 &\textbf{ 0.3535} & \textemdash & 0.2720 & \textbf{0.0337} & \textemdash & \textbf{0.1637} \\
        \( k=7 \) & 0.0705 & \textemdash & 0.0469 & 0.3436 & \textemdash & 0.2516 & 0.0278 & \textemdash & 0.1369 \\
        \( k=10 \) & \textbf{0.1026} & \textemdash & 0.0700 & 0.2340 & \textemdash & 0.1629 & 0.0222 & \textemdash & 0.1097 \\
        \bottomrule
    \end{tabular}}
    \caption{Comparison of performance metrics (M-F1, MAE, and ACC) across Breakfast, MultiTHUMOS and EPIC\_KITCHEN datasets over forecast, imputation and anomaly detection tasks for different few-shot values \( k \) given as in context examples. $k = 0$ indicates Zero Shot. Higher M-F1 and ACC values indicate better performance, while lower MAE values are better. MAE computation is not applicable for anomaly detection. Best metric values are indicated in \textbf{bold}.}
    \label{tab:few_shot_comparison_adjusted}
\end{table}

\subsection{LASTS Few Shot}
\label{sec:few-shot-analysis}

\begin{figure}[t]
    \centering
    \includegraphics[width=\textwidth]{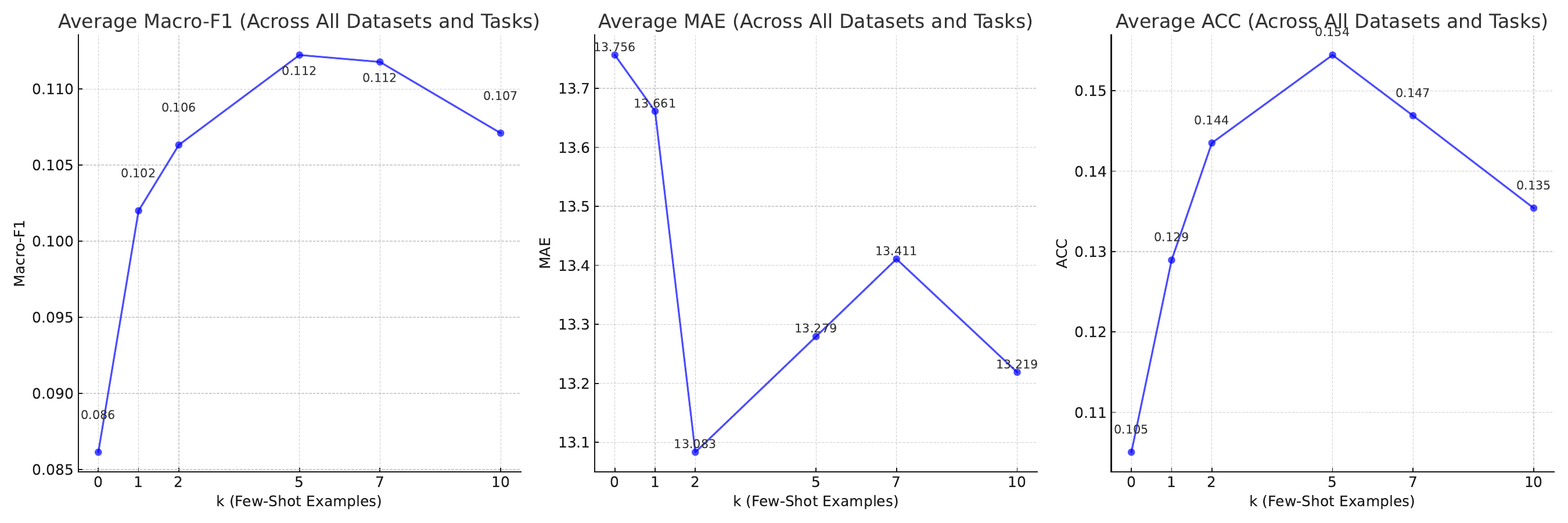}
    \caption{Average values of Macro-F1, MAE, and ACC across all datasets and tasks for different values of \(k\) (number of few-shot examples). Higher values indicate better performance for Macro-F1 and ACC, while lower values indicate better performance for MAE. The results indicate that on an average, $k=5$ works best.}
    \label{fig:performance_vs_k}
\end{figure}

We study the impact of varying the number of examples (\(k\)) in the few-shot setting to determine the optimal value of \(k\) for our method. Specifically, we evaluate the performance of LASTS Few Shot on all datasets and tasks using different \(k\) values, ranging from \(k=0\) (Zero Shot) to \(k=10\). As shown in \autoref{fig:performance_vs_k} and detailed in \autoref{tab:few_shot_comparison_adjusted}, the performance metrics—Macro-F1, MAE, and ACC—improve significantly as \(k\) increases from 0 to 5. However, further increases in \(k\) beyond 5 do not consistently yield improvements and, in some cases, result in marginal performance degradation.

On average, \(k=5\) achieves the best balance across all metrics and datasets. Therefore, we adopt \(k=5\) as the default value for LASTS Few Shot and include it as the entry for "LASTS Few Shot" in \autoref{tab:performance_comparison}. 

\subsection{Further analysis on Stochastic Soft Prompts (StoP)}
\label{sec:stop_analysis}

In this section, we comment on the structure learned by StoP prompts and discuss the practical benefits of Stochastic Soft Prompts.

\begin{figure}[]
    \centering
    \begin{minipage}[t]{0.34\textwidth}
        \centering
        \includegraphics[clip, trim=0cm 0cm 0cm 2.3cm, width=\linewidth]{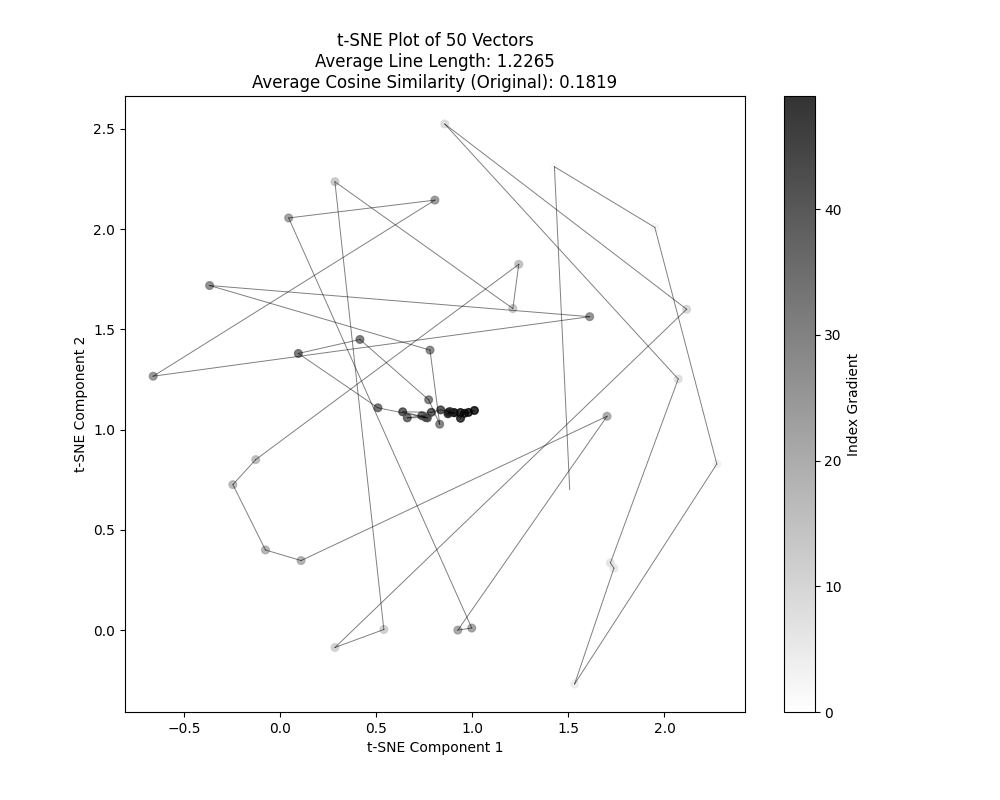}
    \end{minipage}
    \begin{minipage}[t]{0.55\textwidth}
        \centering
        \includegraphics[clip, trim=0cm 0cm 0cm 0cm, width=\linewidth]{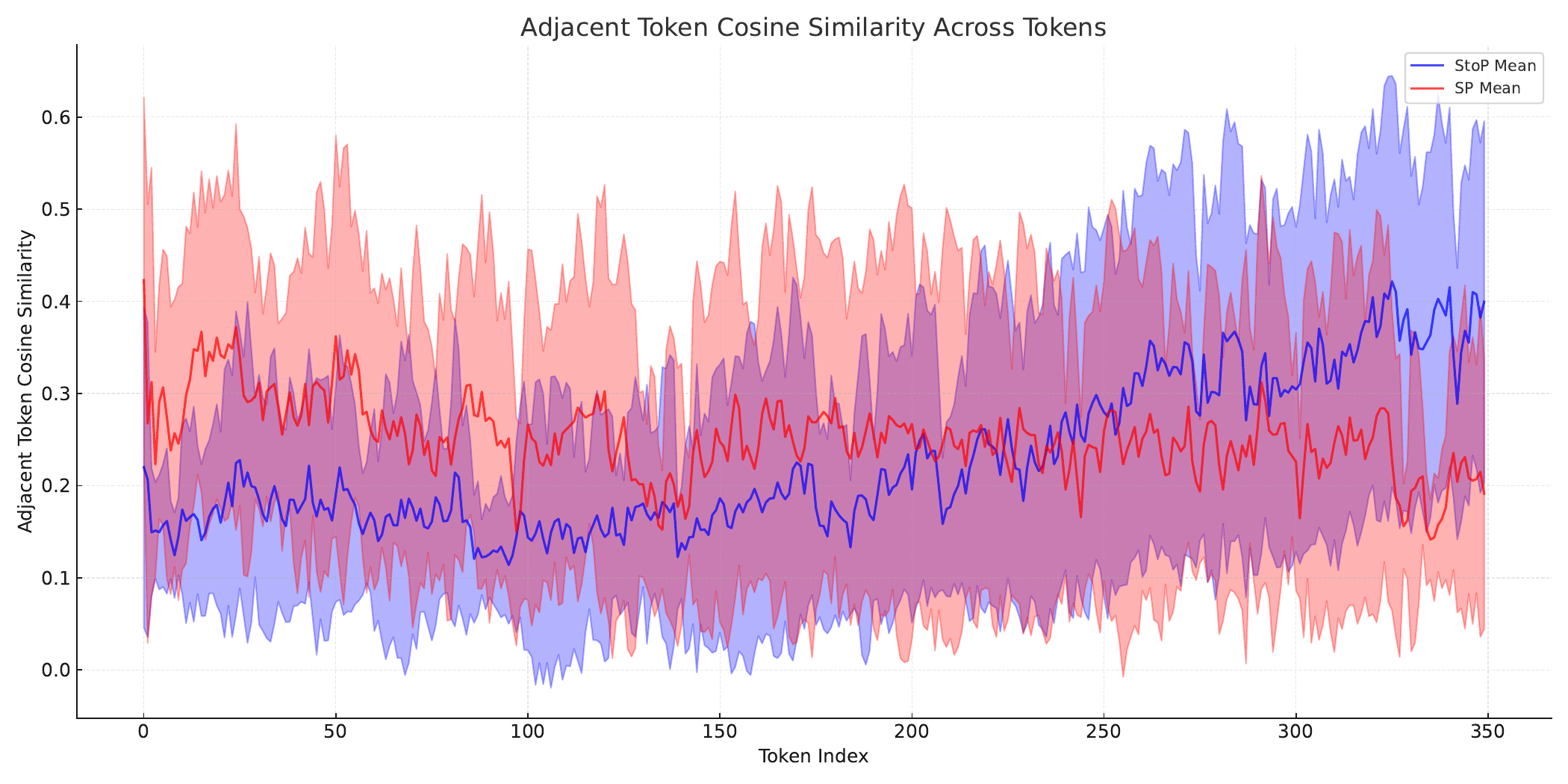}
    \end{minipage}
    
    \caption{\textit{Left}: t-SNE projections of Stochastic Soft Prompt (StoP) tokens with a prompt length of $50$ on the Breakfast dataset for the forecasting task. Adjacent tokens are connected by a line, and the color darkens as the token index increases. The presence of lighter tokens on the periphery and darker tokens in the center indicates that the initial tokens learn very diverse information, while this diversity diminishes as the token index increases. \textit{Right}: Pairwise cosine similarity of the first $350$ tokens of a stochastic soft prompt and a soft prompt learned on the Breakfast dataset for forecasting. We observe that in StoP, the initial cosine similarities are smaller and increase as the token index increases, while no such variation by token index is present in a normal soft prompt.}
    \label{fig:coarse-to-fine}
\end{figure}

\paragraph{Evidence for Coarse-to-Fine Structure}
The prompts learned through Stochastic Soft Prompts (StoP) suggest the presence of a structured coarse-to-fine hierarchy. In this structure, the first few tokens appear to encode broader task-level information, while later tokens may refine predictions by adding more detailed nuances. Below, we provide observations that support this behavior:

\begin{enumerate}
    \item \textbf{t-SNE Projections:} Visualizations of t-SNE projections (see \autoref{fig:coarse-to-fine}) suggest that the first few tokens in StoP prompts may encode more diverse or independent representations, as indicated by their wider spread in the projection space. In contrast, the later tokens tend to cluster more closely together, potentially reflecting the refinement of previously encoded information.

    \item \textbf{Cosine Similarity:} Adjacent tokens at the beginning of the StoP prompt tend to exhibit lower cosine similarity compared to tokens later in the prompt (see \autoref{fig:coarse-to-fine}). This pattern suggests more diverse information being captured at the beginning of the prompt. Standard soft prompts, however, show uniformly high cosine similarities across all tokens, lacking this structure. 

    \item \textbf{Prefix Validity:} \autoref{fig:vallid_porefixes} indicate that any prefix of a StoP prompt serves as a valid standalone prompt, with additional tokens refining the predictions. This behavior suggests that early tokens convey broad task-level information, while later tokens refine and add finer-grained details.
\end{enumerate}

% \begin{figure}[t]
%     \centering
%     % First figure with cropping
%     \begin{subfigure}[b]{0.3\textwidth}
%         \centering
%         \includegraphics[clip, trim=0cm 0cm 0cm 2.2cm, width=\linewidth]{figures/50_hsp_forecast_breakfast.png}
%         \caption{Forecast on Breakfast Dataset}
%         \label{fig:forecast_breakfast}
%     \end{subfigure}
%     \hfill
%     % Second figure
%     \begin{subfigure}[b]{0.6\textwidth}
%         \centering
%         \includegraphics[width=\linewidth]{figures/Final_Adjusted_Cosine_Similarity_StoP_SP.pdf}
%         \caption{Adjusted Cosine Similarity}
%         \label{fig:cosine_similarity}
%     \end{subfigure}
%     \caption{Comparison of Forecasting and Adjusted Cosine Similarity.}
%     \label{fig:side_by_side_figures}
% \end{figure}

\vspace{1in}
\paragraph{Practical Benefits of StoP} We observe that StoP offers many benefits over standard soft prompting:

\begin{enumerate}
    \item \textbf{Improved Generalization:} StoP prompts achieve better generalization compared to standard soft prompts, with an average improvement of \textbf{12.69\%} in Macro-F1 across all datasets (Breakfast, MultiTHUMOS, and EPIC\_KITCHENS) and tasks (Forecast, Imputation, Anomaly Detection) (see \autoref{tab:compare_stop_and_sp})

    \item \textbf{Faster Training:} The stochastic nature of StoP reduces training time by approximately \textbf{25\%}, making it more efficient than standard soft prompting.

    \item \textbf{Resource Efficiency:} StoP enables flexible deployment in resource-constrained environments. Longer trained StoP prompts can be truncated to prefixes as needed, allowing for adaptable inference without compromising performance.
\end{enumerate}

\subsection{Complete Evaluation on Textual Datasets}
\label{sec:accuracy_numbers}
Here we reproduce the main table from our paper, along with accuracy numbers for the interested readers.

\begin{table}[ht]
    \centering
        \caption{\textit{Performance of our models on three textual datasets for forecasting, imputation, and anomaly detection tasks. Metrics are macro F1, and accuracy (ACC) for event type prediction and MAE for event time prediction. The \textbf{best result} in each class is highlighted in bold, and the \underline{second-best result} is underlined. Note that for anomaly detection, since the task involves identifying only the anomalous event, the MAE metric is not applicable and Chronos and LLMProcesses are not adaptable (see  \ref{sec:llms_for_ts_baselines}). A \textsuperscript{*} indicates our method. We use $5$ examples for few shot results (see \ref{sec:few-shot-analysis}). }} 
    \resizebox{\columnwidth}{!}{
    \begin{tabular}{l|ccc|ccc|ccc}
        \toprule
        \textbf{Model} & \multicolumn{3}{c|}{\textbf{Breakfast}} & \multicolumn{3}{c|}{\textbf{MultiTHUMOS}} & \multicolumn{3}{c}{\textbf{EPIC-KITCHENS}} \\
        \midrule
        & M-F1 ($\uparrow$) & MAE ($\downarrow$) & ACC ($\uparrow$) & M-F1 ($\uparrow$) & MAE ($\downarrow$) & ACC ($\uparrow$) & M-F1 ($\uparrow$) & MAE ($\downarrow$) & ACC ($\uparrow$) \\
        \midrule
        & \multicolumn{9}{c}{Forecast}  \\
        \midrule
        Random & 0.0162 & 40.1513 & 0.0201 & 0.0417 & 1.8803 & 0.0382 & 0.0000 & 3.2001 & 0.0001 \\
        Chronos & 0.0011 & 43.0502	& 0.0021 & 0.0265 & 1.9805 & 0.0279 &	0.0000 & 3.5925	& 0.0005\\
        LLMTime & 0.0240 & 37.3902 &	0.0288	& 0.1280 & 2.2060 & 0.1235 & 0.0040 & 4.8948 & 0.0458\\
        LLMProcesses & 0.0337 & 44.9856 &	0.0845 & 0.1278	& 2.0471 &	0.0970	& 0.0049 &	4.3843 & 0.0703\\
        LASTS Zero Shot\textsuperscript{*} & 0.0604 & 38.1630 & 0.0969 & 0.1361 & 1.8868 & 0.1826 & 0.0105 & 3.1566 & 0.0920 \\
        LASTS Few Shot\textsuperscript{*} & 0.1518 & 35.5605 &	0.2133 & 0.1676 & 1.8114 & 0.2581 &	0.0149 & 3.3092 & 0.1150 \\
        LASTS + QLORA\textsuperscript{*} & \underline{0.2558} & 33.9737 & \underline{0.3763} & 0.3218 & 1.7281 & 0.4337 & 0.0764 & \underline{2.8964} & 0.2160 \\
        LASTS + SP\textsuperscript{*} & 0.2341 & \underline{32.8417} & 0.3691 & \underline{0.3707} & \underline{1.6630} & \underline{0.4782} & \underline{0.0780} & \textbf{2.8830} & \underline{0.2217} \\
        LASTS + StoP\textsuperscript{*} & \textbf{0.2633} & \textbf{32.5464} & \textbf{0.3880} & \textbf{0.3947} & \textbf{1.6503} & \textbf{0.4784} & \textbf{0.0797} & 3.0318 & \textbf{0.2298} \\
        \midrule
        & \multicolumn{9}{c}{Imputation}  \\
        % & M-F1 & MAE & ACC & M-F1 & MAE & ACC & M-F1 & MAE & ACC \\
        \midrule
        Random & 0.0168 & 37.7029 & 0.0214 & 0.0435 & 2.3622 & 0.0416 & 0.0000 & 3.4269 & 0.0001 \\
        Chronos & 0.0013 & 38.4039 & 0.0044 & 0.0294 & 2.3971 & 0.0312 &	0.0000 & 3.6955	& 0.0000\\
        LLMTime & 0.0137 & 35.9899	& 0.0381 & 0.0968	& 2.6998 & 0.1330 & 0.0005 &	3.6750 & 0.0314 \\
        LLMProcesses & 0.0156 & 34.7117 & 0.0488 & 0.1123 & 2.3786 & 0.1430 & 0.0008 & 4.2600 & 0.0371\\
        LASTS Zero Shot\textsuperscript{*} & 0.0263 & 33.0097 & 0.0594	& 0.0915 & 2.6696 & 0.1210 & 0.0015 & 3.6527 & 0.0446 \\
        LASTS Few Shot\textsuperscript{*} & 0.0520 & 33.3440 & 0.1001 & 0.1013	& 2.3982 & 0.1569 & 0.0023 & 3.2528	& 0.0547 \\
        LASTS + QLORA\textsuperscript{*} & 0.1688 & \underline{28.5638} & \underline{0.2500} & \underline{0.2132} & \textbf{2.2179} & \underline{0.2744} & 0.0378 & \underline{3.1194} & 0.1137 \\
        LASTS + SP\textsuperscript{*} & \underline{0.1581} & 28.8503 & 0.2264 & 0.2044 & 2.4092 & 0.2528 & \underline{0.0423} & 3.1456 & \underline{0.1270} \\
        LASTS + StoP\textsuperscript{*} & \textbf{0.2064} & \textbf{28.2251} & \textbf{0.2740} & \textbf{0.2213} & \underline{2.3445} & \textbf{0.2839} & \textbf{0.0610} & \textbf{3.1116} & \textbf{0.1424} \\
        \midrule
        & \multicolumn{9}{c}{Anomaly Detection}  \\
        % & M-F1 & MAE & ACC & M-F1 & MAE & ACC & M-F1 & MAE & ACC \\
        \midrule
        Random & 0.0349 & \textemdash & 0.0396 & 0.0381 & \textemdash & 0.0552 & 0.0238 & \textemdash & 0.0307 \\
        LLMTime & 0.0240 & \textemdash & 0.0288 &	0.0415 & \textemdash & 0.0639 & 0.0048	& \textemdash & 0.0650\\
        LASTS Zero Shot\textsuperscript{*} & 0.0923 & \textemdash & 0.0763 & 0.2755 & \textemdash & 0.1949 & 0.0159 & \textemdash & 0.0777 \\
        LASTS Few Shot\textsuperscript{*} & 0.0837 & \textemdash	& 0.0563	& 0.3535 & \textemdash & 0.2720 & 0.0337 & \textemdash & 0.1637 \\
        LASTS + QLORA\textsuperscript{*} & \underline{0.7011} & \textemdash & \underline{0.6478} & \underline{0.6003} & \textemdash & \underline{0.5084} & \underline{0.6520} & \textemdash & \underline{0.6988} \\
        LASTS + SP\textsuperscript{*} & 0.6520 & \textemdash & 0.5937 & 0.5231 & \textemdash & 0.4657 & 0.6159 & \textemdash & 0.6635 \\
        LASTS + StoP\textsuperscript{*} &\textbf{0.7198} & \textemdash & \textbf{0.6698} & \textbf{0.6045} & \textemdash & \textbf{0.5168} & \textbf{0.6603} & \textemdash & \textbf{0.7037}\\
        \bottomrule
    \end{tabular}
     }

    \label{tab:performance_comparison}
\end{table}